\documentclass[10pt,twocolumn,letterpaper]{article}

\usepackage{cvpr} 

\usepackage{bigstrut}
\usepackage{comment}
\usepackage{booktabs}
\usepackage{multirow}
\usepackage{url}
\usepackage{soul}
\usepackage{fontawesome}
\usepackage{subcaption}
\usepackage[font={footnotesize}]{caption}

\usepackage[table,xcdraw,dvipsnames]{xcolor}
\definecolor{gray}{RGB}{170,170,170}
\newcommand{\tsb}[1]{\textsubscript{\textcolor{gray}{$\pm$#1}}}
\newcommand{\boldparagraphstart}[1]{\vspace{2pt}\noindent \textbf{#1}}

\newenvironment{myitem}{\begin{list}{$\bullet$}
{\setlength{\itemsep}{-0pt}
\setlength{\topsep}{0pt}
\setlength{\labelwidth}{5pt}
\setlength{\leftmargin}{10pt}
\setlength{\parsep}{-0pt}
\setlength{\itemsep}{0pt}
\setlength{\partopsep}{0pt}}}%
{\end{list}}
\usepackage{amsmath}

\expandafter\def\expandafter\normalsize\expandafter{%
    \normalsize%
    \setlength\abovedisplayskip{0pt}%
    \setlength\belowdisplayskip{0pt}%
    \setlength\abovedisplayshortskip{3pt}%
    \setlength\belowdisplayshortskip{0pt}%
}

\newcommand*{\affmark}[1][*]{\textsuperscript{#1}}

\definecolor{cvprblue}{rgb}{0.21,0.49,0.74}
\usepackage[pagebackref,breaklinks,colorlinks,citecolor=cvprblue]{hyperref}
\usepackage{siunitx}

\def\paperID{1552} 
\def\confName{CVPR}
\def\confYear{2024}

\title{Neural Implicit Representation for Building 
Digital Twins of\\
Unknown Articulated Objects}

\author{Yijia Weng\affmark[1,2]\thanks{work done during internship at NVIDIA} \quad Bowen Wen\affmark[1] \quad Jonathan Tremblay\affmark[1] \quad Valts Blukis\affmark[1] \\
Dieter Fox\affmark[1] \quad Leonidas Guibas\affmark[2] \quad Stan Birchfield\affmark[1] \quad\\ \\
{\affmark[1]NVIDIA} \quad
{\affmark[2]Stanford University}
}

\begin{document}
\maketitle

\begin{abstract}

We address the problem of building digital twins of unknown articulated objects from two RGBD scans of the object at different articulation states. 
We decompose the problem into two stages, each addressing distinct aspects. Our method first reconstructs object-level shape at each state, then recovers the underlying articulation model including part segmentation and joint articulations that associate the two states. 
By explicitly modeling point-level correspondences and exploiting cues from images, 3D reconstructions, and kinematics, our method yields more accurate and stable results compared to prior work. 
It also handles more than one movable part and does not rely on any object shape or structure priors.
Project page: \url{https://github.com/NVlabs/DigitalTwinArt}

\end{abstract}    
\vspace{-10pt}
\section{Introduction}
\label{sec:intro}
\begin{figure}[t]
  \centering
  \includegraphics[width=\linewidth]{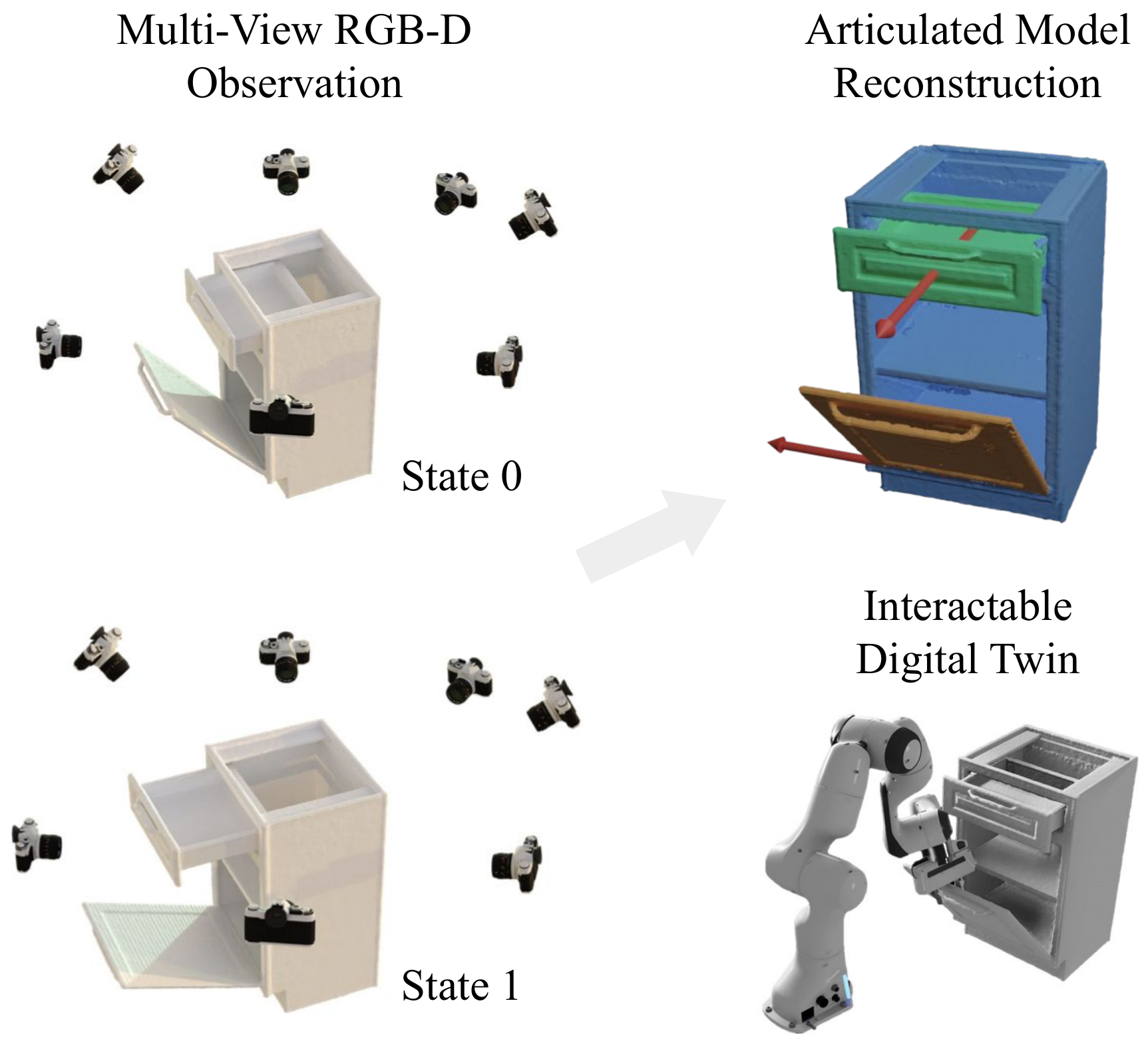}
  \vspace{-25pt}
  \caption{Our method requires two RGB-D scans of the object in each of two articulation states (left).  The output is a 3D reconstruction with parts segmented, joint types identified, and joint axes estimated (top right).
  Note that multiple joints are allowed. 
 The resulting digital twin can be imported into a physics-based simulator for interaction (bottom right).}
  \label{fig:intro}
  \vspace{-15pt}
\end{figure}

Articulated objects are all around us. Whenever we open a door, close a drawer, turn on a faucet, or use scissors, we leverage the complex, physics-based understanding of various object parts and how they interact. Reconstructing novel articulated objects from visual observations is therefore an important problem for robotics and mixed reality. In this work, we aim to democratize the process of building a 3D reconstruction that accurately describes an articulated object, including the part geometries, segmentation and their joint articulations, as shown in Figure~\ref{fig:intro}.

The problem of generating digital twins of articulated objects has been long studied~\cite{huang2014ras:art,hu2017learning,wang2019shape2motion,yi2018deep, zeng2021visual,heppert2022category,mu2021sdf,jiang2022ditto,heppert2023carto,kawana2022unsupervised,wei2022self}. Two recent approaches to this problem are Ditto~\cite{jiang2022ditto} and PARIS~\cite{liu2023paris}. Both works reconstruct part-level geometries and the articulation model based on observations of the object at two joint states.
Ditto is a feed-forward method that takes two multi-view fused point clouds as input. It is trained on a collection of objects from certain categories. Although Ditto shows generalizability to objects unseen during training, it is not equipped with the capability to handle arbitrary unknown objects drastically different from the training categories.
PARIS takes multi-view posed images as input and optimizes an implicit representation directly from the input data without pretraining, ensuring better generalizability. However, the optimization process of PARIS depends significantly on initializations and lacks stability, as we will show in the experimental results. In addition, both PARIS and Ditto only handle two-part objects. 

In this paper, we take a step toward addressing the limitations of prior works by proposing a system with the following unique characteristics: a) the ability to handle arbitrary novel objects, regardless of the discrepancy of their motion, shape, or size from the training distribution; b) the scalability to objects with multiple moving parts; c) the robustness to initializations of the high dimensional optimization space of the articulation reconstruction problem.

Given multi-view RGB-D observations of the object at two different joint states, our proposed method reconstructs articulated part meshes and joint parameters. We adopt a two-stage approach, where
we first reconstruct the object at each given state with an SDF representation, and then estimate the articulation model including part segmentation and joint parameters. We explicitly derive point-level correspondences between the two states from the articulation model, which can be readily supervised by minimizing the following losses: 
1) consistency between 3D local geometry from one state to the other, 
2) 2D pixel correspondences from image feature matches, 
and, 3) physically-based reasoning in the form of minimizing articulation collisions.

We demonstrate the efficacy of our method on multiple challenging datasets, such as 
the dataset introduced by PARIS~\cite{liu2023paris} which includes both synthetic and real scenes, 
We also introduce a novel synthetic dataset composed by objects with more than one joint. 
Extensive experiments indicate our approach generalizes to various types of objects, 
including those challenging ones consisting of both revolute and prismatic joints. 
Our method is also shown to produce more stable results than baselines under different initializations. We summarize our contributions as follows:

\begin{myitem}
    \item We present a framework that reconstructs the geometry and articulation model of unknown articulated objects. It is per-object optimized and applicable to arbitrary articulated objects without assuming any object shape or structure priors.
    \item Our method decouples the problem into object shape reconstruction and articulation model reasoning. By jointly optimizing a set of loss terms on a \emph{point correspondence field}, derived from the articulation model, we effectively leverage cues from image feature matching, 3D geometry reconstructions, as well as kinematic rules.
    \item Extensive evaluation on both synthetic and real-world data indicates our approach outperforms existing state-of-art methods consistently and stably.
    \item  We demonstrate generalizability to complex unknown articulated objects consisting of more than one movable part, using only multi-view scans at two different articulation states.
\end{myitem}

\section{Related Work}
\label{sec:related}


\boldparagraphstart{Articulated Object Prior Learning.}
A number of works leverage deep learning to train over large-scale 3D articulated assets offline in order to learn articulation priors, including part segmentation~\cite{jiang2022opd,abdul2022learning,hu2017learning,wang2019shape2motion,shi2021self,tseng2022cla,huang2021multibodysync,yi2018deep}, kinematic structure~\cite{yi2018deep,abdul2022learning,wang2019shape2motion,hu2017learning,zeng2021visual,heppert2022category,shi2021self,wei2022self}, pose estimation~\cite{weng2021captra,li2020category,kawana2022unsupervised,heppert2022category,tseng2022cla}, and articulated shape reconstruction~\cite{mu2021sdf,jiang2022ditto,heppert2023carto,kawana2022unsupervised,wei2022self}. In particular, Ditto~\cite{jiang2022ditto} and CARTO~\cite{heppert2023carto} share the same objective as ours in building a full digital twin of the object, including shape reconstruction, part segmentation and articulation reasoning. Ditto~\cite{jiang2022ditto} builds on top of PointNet++~\cite{qi2017pointnet++} to process multi-view fused point cloud. CARTO~\cite{heppert2023carto} learns a single geometry and articulation decoder for multiple object categories by taking as input the stereo images. While promising results have been shown by the above methods, collecting large amount of training data is non-trivial in the real world due to the annotation complexity. The availability of 3D articulated object models is also notably limited compared to their single rigid counterparts to produce diverse synthetic training data. This results in struggling  with out-of-distribution test set, as validated in our experiments. In contrast, our method does not require training on articulated assets and can be applied to a broad range of unknown articulated objects without any category restriction.

\boldparagraphstart{Per-Object Optimization.} Per-object optimization methods perform test-time optimization to better adapt to the novel unknown objects~\cite{noguchi2022watch,liu2023paris,liu2023building}. 
By circumventing learning priors on 3D articulated assets, this type of approach can in theory generalize to arbitrarily unknown objects. Watch-it-move~\cite{noguchi2022watch} demonstrates self-discovery of 3D joints for novel view synthesis and re-posing. However, it focuses on revolute joints and objects such as humans, quadrupeds and robotic arms as opposed to the daily-life objects considered here. \cite{liu2023building} proposes an energy minimization approach to jointly optimize the part segmentation, transformation, and kinematics, while requiring a sequence of complete point cloud as input. Among these methods, PARIS~\cite{liu2023paris} shares the closest setup considered in this work by taking two scans corresponding to the initial and final states of the unknown object and building a full digital twin. It focuses on objects with a single movable part, modeling both the static and dynamic part separately with individual neural fields.  As we show, this design decision results in less robustness and efficiency, and it thus prevents generalization to more complex multi-joint objects, such as those handled by our method.

\boldparagraphstart{Articulation Reasoning by Interaction.}  Prior work~\cite{hsu2023ditto,gadre2021act,ma2023sim2real} leverages physical interaction to create novel sensory information so as to reason the articulation model based on object state changes. \cite{katz2013interactive,huang2014ras:art} pioneer the effort to introduce interactive perception into the estimation of articulation models. Follow up works further explore with hierarchical recursive Bayesian filters~\cite{martin2016integrated}, probabilistic models~\cite{sturm2011probabilistic}, geometric models from multi-view stereo~\cite{huang2014ras:art}, and feature tracking~\cite{pillai2015learning}. Where2Act~\cite{mo2021where2act} presents a learnable framework
to estimate action affordance on articulated objects from
a single RGB image or point cloud while limited to single step interaction. AtP~\cite{gadre2021act}  learns interaction strategies to isolate parts for effective part segmentation and kinematic reasoning.  However, most of the methods focus on learning interaction policies for effective part segmentation or motion analysis and do not aim for 3D reconstruction, which is part of the goal in this work. Recent work \cite{hsu2023ditto} extends Ditto~\cite{jiang2022ditto} to an interactive setup which enables full digital twining. Nevertheless, its dependency on pretraining shares the similar issues as the articulation prior learning methods. The assumption on perfect depth sensing without viewpoint issues also hinders direct application to noisy real-world data.

\section{Method}

\begin{figure*}[t]
  \centering
  \includegraphics[width=0.99\linewidth]{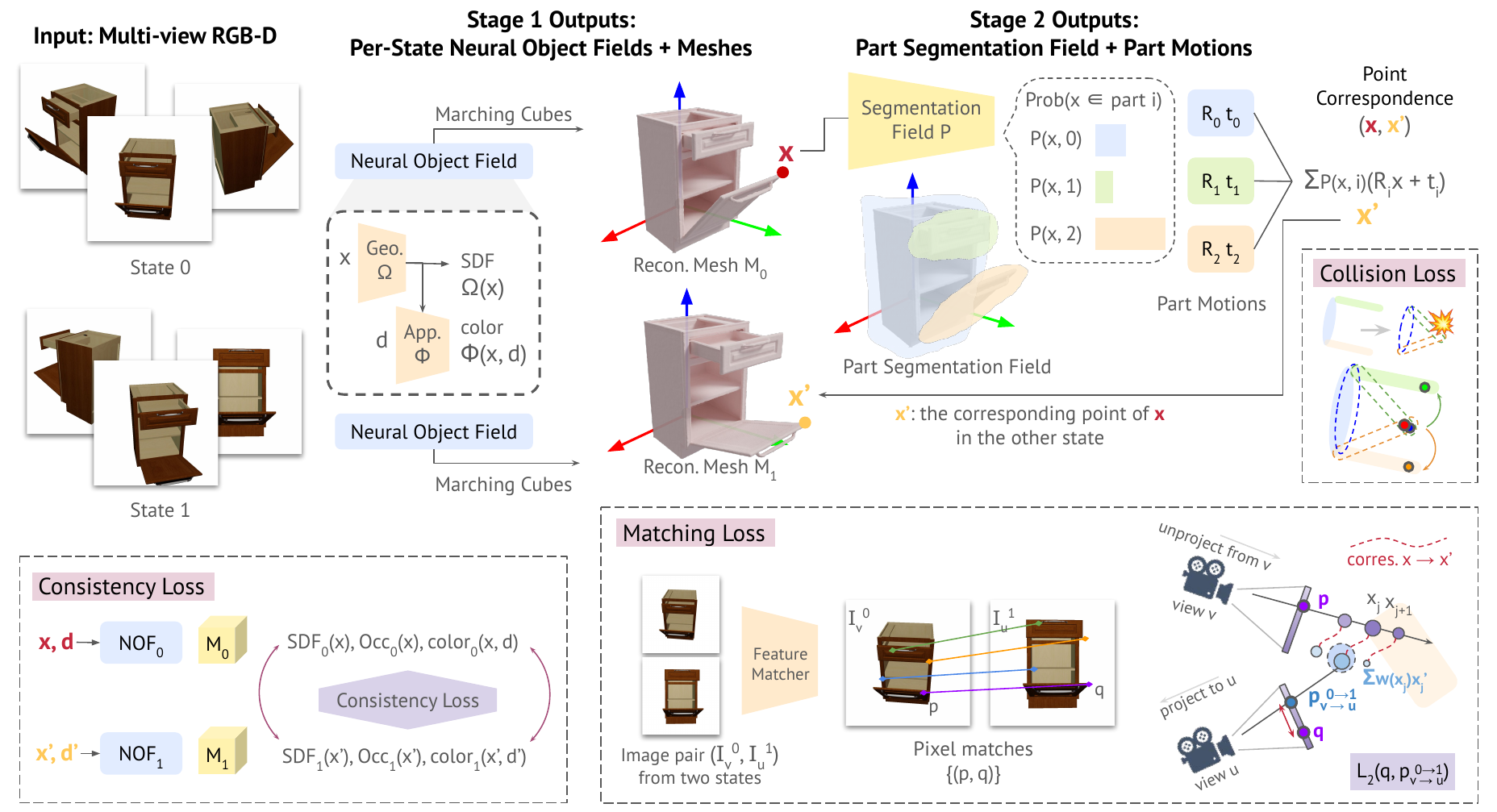}
  \vspace{-10pt}
  \caption{Overview of our method.  In Stage 1, given multi-view RGB-D scans for the object at the initial and final articulation states, two neural object fields are optimized for each state. Upon learning convergence, the meshes corresponding to the two states are extracted.  In Stage 2, the part segmentation field and per-part motions are optimized with three losses: consistency, matching, and collision. Together, the segmentation field and part motions yield point correspondence between the two states.}
  \label{fig:overview}
  \vspace{-15pt}
\end{figure*}

We address the problem of building digital twins of unknown multi-part articulated objects from observations of the object at two different joint states.
Specifically, we reconstruct per-part shapes and the articulation model of the object, given multi-view RGB-D observations and object masks $\{(I^t_v, \text{Depth}^t_v, \text{Mask}^t_v)\}_{v = 0, \ldots, V - 1}$ with known camera parameters at object initial state $t = 0$ and final state $t=1$.
Typically the number of images $V \approx 100$. We also assume the number of joints is given.

Figure \ref{fig:overview} presents an overview of our framework. We factorize the reconstruction problem into two stages with distinct focuses.
Stage one (\S\ref{sec:method_stage_1}) reconstructs the object-level shape at each state, which are independent of articulation. 
Stage two (\S\ref{sec:method_stage_2}) recovers the articulation model including part segmentation and part motions by exploiting correspondences between per-state reconstructions. 

\subsection{Per-State Object Reconstruction}\label{sec:method_stage_1}

Given multi-view posed RGB-D images of the object $\mathcal{O}^t$ at state $t \in \{0, 1\}$, we aim to reconstruct object geometry, represented by a Neural Object Field \cite{wen2023bundlesdf} $(\Omega^t, \Phi^t)$ (we omit $t$ for simplicity in the following), where the geometry network $\Omega: \mathbf{x} \mapsto d$ maps spatial point $\mathbf{x} \in \mathbb{R}^3$ to its truncated signed distance $d \in \mathbb{R}$, and the appearance network $\Phi: (\mathbf{x}, \mathbf{d}) \mapsto \mathbf{c}$ maps point $\mathbf{x} \in \mathbb{R}^3$ and view direction $\mathbf{d} \in \mathbb{S}^2$ to RGB color $\mathbf{c} \in \mathbb{R}^3_{+}$. 

The networks $\Omega$ and $\Phi$ are implemented with multi-resolution hash encoding~\cite{muller2022instant} and are supervised with RGB-D images via color rendering loss $\mathcal{L}_c$ and SDF loss $\mathcal{L}_{SDF}$. We follow the approach of BundleSDF~\cite{wen2023bundlesdf} and defer details to the appendix.

After optimization, we obtain the object mesh $\mathcal{M}^t$  by extracting the zero level set from $\Omega$ using marching cubes~\cite{lorensen1998marching}, from which we can further compute the Euclidean signed distance field (ESDF) $\tilde{\Omega}(\mathbf{x})$, as well as the occupancy field $\operatorname{Occ}(\mathbf{x})$, defined as
\begin{align}
\operatorname{Occ}(\mathbf{x}) = \operatorname{clip}(0.5 - \frac{\tilde{\Omega}(\mathbf{x})}{s}, 0, 1),\label{eq:occupancy}
\end{align}
where $s$ is set to a small number to make the function transition continuously near the object surface.


\subsection{Segmentation and Motion Reconstruction}\label{sec:method_stage_2}

Given object-level reconstructions $(\mathcal{M}^t, \tilde{\Omega}^t, \operatorname{Occ}^t, \Phi^t)$ at two different articulation states $t \in \{0, 1\}$, we aim to discover the underlying articulation model that associates them to each other, namely part segmentation and per-part rigid transformations between states. Our key idea is to derive a \textit{point correspondence field} between states from the articulation model and supervise it with rich geometry and appearance information obtained from the first stage. 

For an articulated object with $M$ parts, we model its articulation from state $t$ to state $t' = 1 - t$ with 1) a part segmentation field $f^t: \mathbf{x} \mapsto i$ that maps spatial point $\mathbf{x} \in \mathcal{O}^t$ from the object at state $t$ to a part label $i \in \{0, \ldots, M - 1\}$, and 2) per-part rigid transformation $\mathcal{T}^t_{i} = (R^t_i, \mathbf{t}^t_i) \in \mathbb{SE}(3)$ that transforms part $i$ from state $t$ to state $t'$. 

For differentiable optimization, instead of hard assignment $f$ of points to parts, we model part segmentation as a probability distribution over parts.  Formally, we let $P^t(\mathbf{x}, i)$ be the probability that point $\mathbf{x}$ in state $t$ belongs to part $i$.  

$P^t$ is implemented as a dense voxel-based 3D feature volume followed by MLP segmentation heads. For rigid transformations, we parameterize rotations with the 6D representation as in \cite{zhou2019continuity} and translations as 3D vectors.
We can now derive the \textit{point correspondence field} that maps any object point $\mathbf{x}$ from state $t$ to its new position $\mathbf{x}^{t\to t'}$ at state $t'$ when it moves \textit{forward} with the motion of the part it belongs to. The field can be seen as a way to ``render'' the articulation model for supervision. Formally,
\begin{align}
\mathbf{x}^{t\to t'} = \overrightarrow{\operatorname{Fwd}}(\mathbf{x}, f^t, \mathcal{T}^t) = \sum_{i} P^t(\mathbf{x}, i) (R_{i}^t\mathbf{x} + \mathbf{t}_{i}^t). \label{eq:point_corr_con}
\end{align}
This scheme is similar to classical linear blend skinning~\cite{kavan2007skinning}.

\boldparagraphstart{Shared Motion.} We optimize two articulation models $(f^0, \mathcal{T}^0)$, $(f^1, \mathcal{T}^1)$ starting from both states. As they describe the same articulations, we share the part motions $\mathcal{T}$ to reduce redundancy and share supervision signal. Formally, 
\begin{align}
\mathcal{T}^1_i = (R^0_i, \mathbf{t}^0_i)^{-1} ~,\forall i
\end{align}

Given the \textit{point correspondence field}, we can supervise it with rich geometry and appearance information from object-level reconstructions and image observations. Specifically, we propose the following losses.

\boldparagraphstart{Consistency Loss.} Corresponding points should have consistent local geometry and appearance at their respective states, which we can query from stage one's reconstructions. For near-surface points $\mathbf{x} \in \mathcal{X}^t_{sur\!f} = \{\mathbf{x} \mid  |\tilde{\Omega}(\mathbf{x})| < \lambda_{sur\!f}\}$, we expect its correspondence $\mathbf{x}^{t\to t'}$ to have consistent SDF and color. Formally, we define \textit{SDF consistency loss} $l_{s}$ and \textit{RGB consistency loss} $l_{c}$ as
\begin{align}
&l_{s}(\mathbf{x}) = (\tilde{\Omega}^t(\mathbf{x}) - \tilde{\Omega}^{t'}(\mathbf{x}^{t\to{t'}}))^2, \\
&l_{c}(\mathbf{x}) = \left \|\Phi^t(\mathbf{x}, \mathbf{d}) - \Phi^{t'}(\mathbf{x}^{t\to t'}, \mathbf{d'})\right\|_2^2,
\end{align} 
where $\mathbf{d}$ denotes the direction of the ray $\mathbf{x}$ sampled from, $\mathbf{d'}$ denotes $\mathbf{d}$ transformed by $\mathbf{x}$'s part motion. 

To extend supervision to points away from the surface, for which we are less confident about the reconstructed SDF or color, we enforce consistency on their occupancy values. Formally, we define \textit{occupancy consistency loss} $l_{o}$ as
\begin{align}
l_{o} = \left \|\operatorname{Occ}^t(\mathbf{x}) - \operatorname{Occ}^{t'}(\mathbf{x}^{t\to t'})\right\|_2^2
\end{align}

We enforce SDF and color consistency loss on points $\textbf{x}$ sampled along camera rays $\mathbf{r}(t) = \mathbf{o} + t\mathbf{d}$ and weigh points based on their proximity to the object surface, similar to~\cite{wen2023bundlesdf}. Meanwhile, we enforce occupancy consistency loss on points uniformly sampled from the unit space. Formally, we define consistency loss $\mathcal{L}_{\text{cns}}$ as 
\begin{align}
\mathcal{L}_{\text{cns}} &= \mathbb{E}_{\mathbf{x} \in \mathcal{X}^{t}_{sur\!f}} \left[w^t(\mathbf{x}) (\lambda_s l_s(\mathbf{x})
+ \lambda_c l_c (\mathbf{x}))\right] \notag\\
& + \mathbb{E}_{\mathbf{x}}\left[\lambda_{o}l_{o}(\mathbf{x})\right] \label{eq:consistency_loss}, \\
w(\mathbf{x}) &= \operatorname{Sigmoid}(-\alpha\tilde{\Omega}(\mathbf{x}))\cdot\operatorname{Sigmoid}(\alpha\tilde{\Omega}(\mathbf{x})) \label{eq:render_weight}
\end{align}
where $w(\mathbf{x})$ is a bell-shaped function that peaks at the object surface, hyperparameter $\alpha$ controls its sharpness, and hyperparameters $\lambda_s,\lambda_c,\lambda_o$ weigh different loss terms. 
However, as consistency loss is based on local descriptions, and there is a large solution space for each point, it can sometimes be challenging to arrive at the correct solution when optimizing for consistency loss alone.

\boldparagraphstart{Matching Loss.}  We propose to exploit visual cues from image observations, by leveraging 2D pixel matches across images at two states, obtained by LoFTR~\cite{sun2021loftr}. 

For image $I_{v}^t$ taken from view $v$ at state $t$, we select $K$ images $\{I_{u}^{t'} \mid u \in \mathcal{N}_v\}$ from state $t'$, where $\mathcal{N}_v$ are the viewpoints at $t'$ that are closest to view $v$. 
We feed each image pair $(I_v^t, I_u^{t'})$ to LoFTR to get $L$ pairs of sparse and potentially noisy pixel matches 
$\mathcal{M}_{v, u, t} = \{(\mathbf{p}_j, \mathbf{q}_j)\}_j$. 

For pixel pair $(\mathbf{p}, \mathbf{q})$, let $\mathbf{r}$ be the camera ray from view $v$ that passes through $\mathbf{p}$, the 2D correspondence of $\mathbf{p}$ at state $t'$ from view $u$ can be approximated with 
\begin{align}
\mathbf{p}^{t \to t'}_{v\to u} = \pi_{u}\left(\frac{\sum_{\mathbf{x} \in \mathbf{r}\cap \mathcal{X}_{sur\!f}} w^t(\mathbf{x}) \mathbf{x}^{t \to t'}}{\sum_{\mathbf{x} \in \mathbf{r}\cap \mathcal{X}_{sur\!f}} w^t(\mathbf{x})}\right),
\end{align} 
where $\pi_{u}$ is a projection to view $u$, $w^t(x)$ is as in Eq.~\eqref{eq:render_weight}.

The matching loss is averaged over all matching pixel pairs from all image pairs:
\begin{align}
\mathcal{L}_{\text{match}} = \mathbb{E}_{\substack{(\mathbf{p},\mathbf{q}) \in \mathcal{M}_{v, u, t}, u\in \mathcal{N}_v, \\ v = 0,\ldots, V - 1,  t\in\{0, 1\}}} \left\|\mathbf{p}^{t \to t'}_{v \to u} - \mathbf{q}\right\|_2^2, 
\end{align}
\begin{figure}[t]
  \centering
  \includegraphics[width=0.9\linewidth]{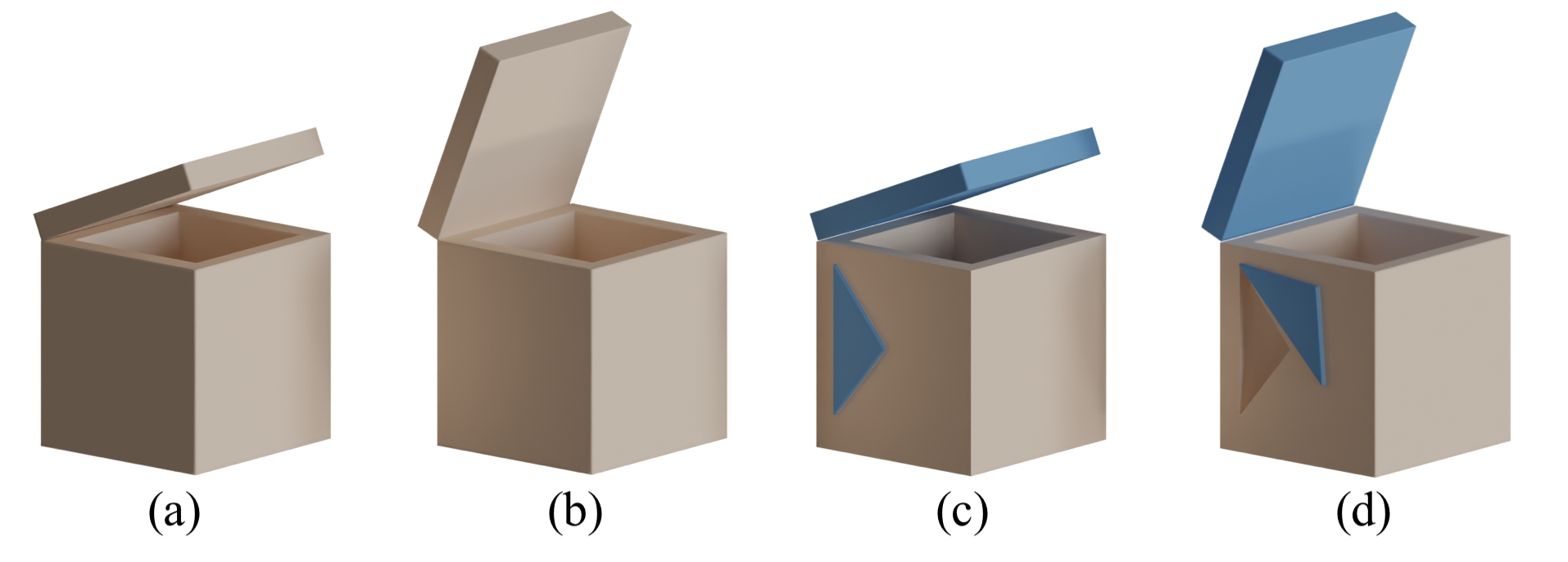}
  \vspace{-10pt}
  \caption{Motivation for collision loss. (a), (b) are the observations for the object at initial and final state respectively. Suppose the object is wrongly segmented as shown in (c), where blue represents the movable part. Moving the part with the forward motion will result in (d). In this case, wrong segmentation field still results in low consistency loss for SDF and color. Therefore, we introduce additional collision loss.
  }
  \label{fig:corner_case}
  \vspace{-10pt}
\end{figure}

\begin{figure}[t]
  \centering
  \includegraphics[width=\linewidth]{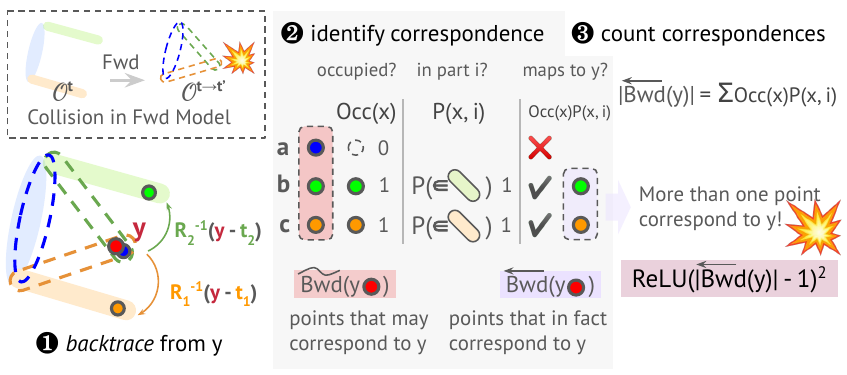}
  \vspace{-20pt}
  \caption{Illustration of the collision loss. 
  We aim to detect and penalize collisions between parts after applying the predicted forward motion (moving the two sticks inwards).
  For point $\mathbf{y}$ at state $t'$, we backtrace a set of points $\widetilde{\operatorname{Bwd}}(\mathbf{y})$ ($\{a, b, c\}$) that may move to $\mathbf{y}$, by transforming $\mathbf{y}$ with each part's inverse motion (moving outwards following the arrows). We then check if the candidate point $\mathbf{x}_i$ obtained with part $i$'s motion is indeed a point in part $i$, by looking up its occupancy and part label. Finally, we obtain the set of points $\overleftarrow{\operatorname{Bwd}}(\mathbf{y})$($\{b, c\}$) that in fact map to $\mathbf{y}$ under the articulation model, and report collision if there are more than one point that maps to $\mathbf{y}$, i.e., $|\overleftarrow{\operatorname{Bwd}}(\mathbf{y})| > 1$.}
  \label{fig:collision_loss}
  \vspace{-18pt}
\end{figure}

\boldparagraphstart{Collision Loss.} A solution that minimizes the consistency loss may still be wrong. As illustrated in Fig.~\ref{fig:corner_case}, a wrong segmentation still leads to a low consistency loss. 
The matching loss will not completely resolve the issue, either, because pixel matches can be noisy and sparse, and they mostly constrain near-surface points and do not work for points deep inside the object. On the other hand, if we look at the per-part transformed object $\mathcal{O}^{t\to t'} = \{\mathbf{x}^{t \to t'} = \overrightarrow{\operatorname{Fwd}}(\mathbf{x}, f^t, \mathcal{T}^t)\}_{\mathbf{x} \in \mathcal{O}^t}$, as shown in Fig.~\ref{fig:corner_case}(d), we do observe artifacts as a result of the wrong segmentation, namely the collision between the triangle and the base. Therefore, we propose to look at the entirety of $\mathcal{O}^{t\to t'}$ and check for artifacts. Fig.~\ref{fig:collision_loss} illustrates the idea. To detect collision, we start from a point $\mathbf{y}$ at state $t'$, and \textit{backtrace} a set of points at state $t$ that may \textit{forward} to it given $(f^t, \mathcal{T}^t)$,
\begin{align} 
\overleftarrow{\operatorname{Bwd}}(\mathbf{y}, f^t, \mathcal{T}^t)  = \{\mathbf{x} \mid \overrightarrow{\operatorname{Fwd}}(\mathbf{x}, f^t, \mathcal{T}^t) = \mathbf{y}\}
\end{align}

To simplify, we consider cases where $\mathbf{x} \in \overleftarrow{\operatorname{Bwd}}(\mathbf{y})$ follows one of $M$ rigid part motions. We can iterate over all possible parts and obtain a candidate set $\widetilde{\operatorname{Bwd}}(\mathbf{y})$,
\begin{align}
\overleftarrow{\operatorname{Bwd}}(\mathbf{y}) &\subset \widetilde{\operatorname{Bwd}}(\mathbf{y}) \notag = \{(R^t_{i})^{-1}(\mathbf{y} - \mathbf{t}^t_i)\}_{i}
\end{align}
During training, we use $\widetilde{\operatorname{Bwd}}(\mathbf{y})$ as an approximation.

Candidate point $\mathbf{x}_i = (R^t_{i})^{-1}(\mathbf{y} - \mathbf{t}^t_i)$ corresponds to $\mathbf{y}$ only if $\mathbf{x}_i$ is on part $i$, which can be verified by checking occupancy $\operatorname{Occ}(\mathbf{x})$ and part segmentation $P(\mathbf{x}, i)$. Formally, we write $\mathbf{x}_i$'s probability of corresponding to $\mathbf{y}$ as 
\begin{align}
a_i = P^t(\mathbf{x}_i, i) \cdot \operatorname{Occ}^t(\mathbf{x}_i),
\end{align}
where $\operatorname{Occ}(\mathbf{x})$ is defined by Eq.~\eqref{eq:occupancy}.

We count the number of points that correspond to $\mathbf{y}$ by summing contributions from all $\mathbf{x}_i$ and report collision when the result is larger than 1. Formally, we define \textit{collision loss} $\mathcal{L}_{\text{coll}}$ as
\begin{align}
&\mathcal{L}_{\text{coll}} = \mathbb{E}_{\mathbf{y} 
}
\left[\operatorname{ReLU}(|\overleftarrow{\operatorname{Bwd}}(\mathbf{y})| - 1)^2\right],\\
\end{align}
where $\mathbf{y}$ is uniformly sampled in the unit space.

\boldparagraphstart{Handling Partial Observation.}
In many cases, we can only observe part of the object due to limited viewpoints or self-occlusions. This may lead to hallucinated regions in object reconstructions that interfere with correspondence reasoning. Moreover, the visible portion of the movable parts vary between states, e.g., an open drawer versus a fully closed one. Points only visible in one state, e.g., those in the interior of the drawer, may not find corresponding points in the other state's reconstruction. To address this issue, we compute the visibility of point $\mathbf{x}$ by projecting it to all camera views and checking if it is in front of the depth (at the projected pixel) beyond a certain threshold $\epsilon$. Formally,
\begin{align}
\operatorname{vis}(\mathbf{x}) = \bigvee_{v=0}^{V - 1} [d_v(\pi_v(\mathbf{x})) + \epsilon > \operatorname{dist}_{v}(\mathbf{x})],\label{eq:visibility}
\end{align}
where $\bigvee$ denotes logical OR, $d_v$ denotes observed depth at view $v$; $\pi_v(\mathbf{x})$ denotes 2D projection; $\operatorname{dist}_v(\mathbf{x})$ denotes the distance along the optical axis from $\mathbf{x}$ to camera origin.  

Let $\mathcal{U}^{t} = \{\mathbf{x} \mid \lnot \operatorname{vis}(\mathbf{x})\}$ denote the set of unobserved points at state $t$. During mesh extraction at the first stage, we enforce the space to be empty at these points by setting their TSDF to $1$, such that surface reconstructions only contain observed regions. We also discount the point consistency loss at $\mathbf{x}$ by a factor of $w_{vis}$ if $\mathbf{x}^{t\to t'}$(Eq. \ref{eq:point_corr_con})~$\in \mathcal{U}^{t'}$, 
i.e., the predicted correspondence in the other state is not observed. $w_{vis}$ is set to a small nonzero number to avoid learning collapse, i.e., making all points correspond to unobserved points to reduce consistency loss. 

Our total loss for the second stage is defined as 
\begin{align}
\mathcal{L} = \lambda_{\text{cns}} \mathcal{L}_{\text{cns}} + \lambda_{\text{match}} \mathcal{L}_{\text{match}} + \lambda_{\text{coll}} \mathcal{L}_{\text{coll}} \label{eq:total_loss}
\end{align}

\boldparagraphstart{Explicit Articulated Object Extraction} Given reconstructed shape and articulation models $(\mathcal{M}^t, P^t, \mathcal{T}^t), t\in\{0,1\}$, we can extract an explicit articulated object model. To predict joint $i$, we take the shared part motion $\mathcal{T}^0_{i} = (R^0_i, \mathbf{t}^0_i)$ and classify joint $i$ as prismatic if $|\operatorname{angle}(R^0_i)| < \tau_{r}$, and revolute otherwise. We then project $\mathcal{T}^0_i$ to the manifold of pure-rotational or translational transformations and compute joint axes and relative joint states. For part-level geometry, we first identify the state $t^*\in \{0, 1\}$ with better part visibility, e.g. when the drawer is open instead of closed. We then compute hard segmentation $f^{t^{*}}(\mathbf{x}) = \arg\max_{i} P^{t^{*}}(\mathbf{x}, i)$, and extract each part mesh as $\mathcal{P}_i^{t^*} = \{\mathbf{v} \mid \mathbf{v} \in \mathcal{M}^{t^*}, f^{t^{*}}(\mathbf{v}) = i\}$.

\section{Experiments}

\begin{figure*}[t]
  \centering
  \includegraphics[width=0.9\linewidth]{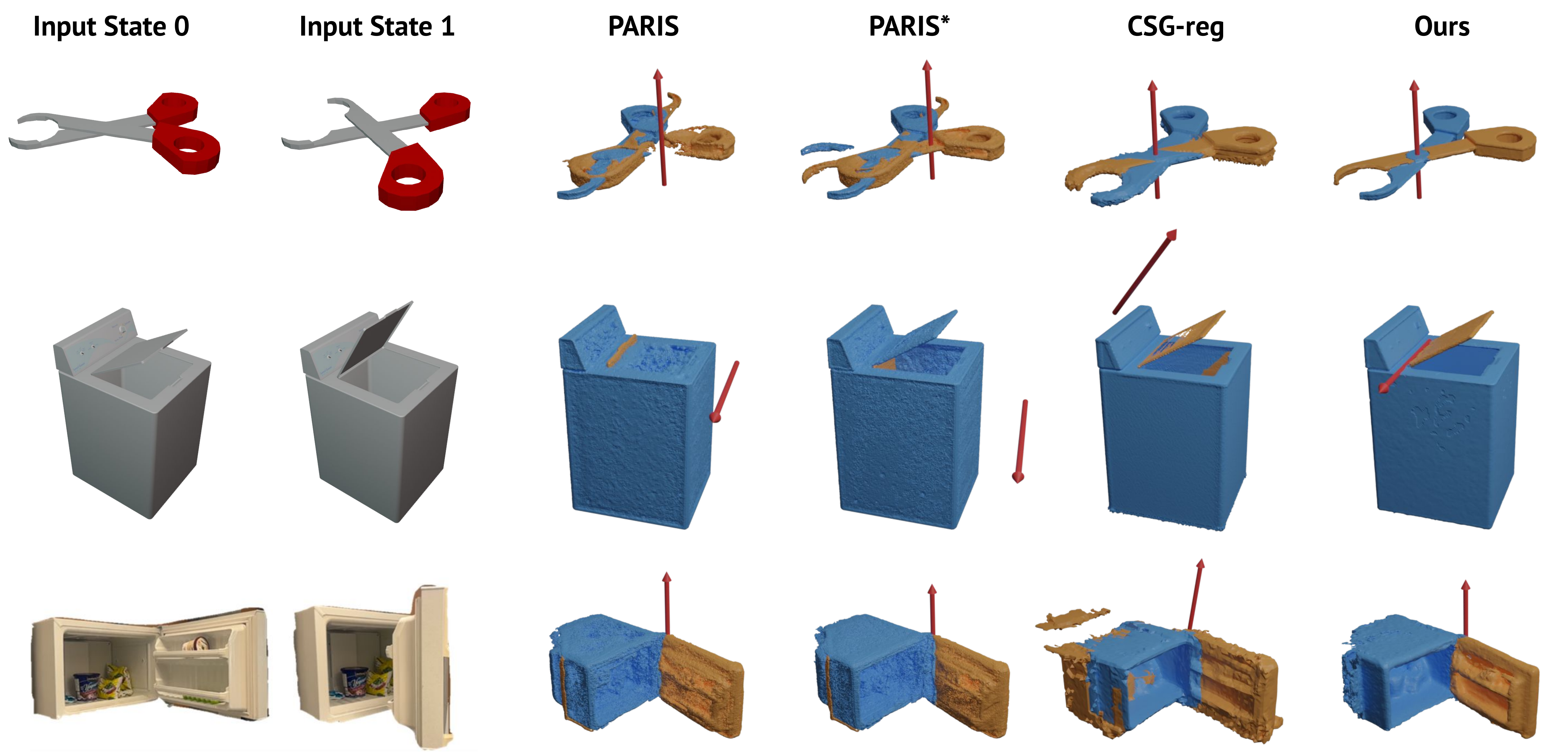}
  \vspace{-10pt}
  \caption{Qualitative results of shape reconstruction,  part segmentation and joint prediction on PARIS dataset~\cite{liu2023paris}.  The top two rows correspond to synthetic data. The bottom row corresponds to real-world data. While PARIS and PARIS* occasionally work for these objects, depending upon the random seed, they often fail. Shown are the results from a typical trial that achieves near-average results.}
  \label{fig:paris_syn}
  \vspace{-0pt}
\end{figure*}

\begin{figure}[t]
  \centering
  \includegraphics[width=1.0\linewidth]{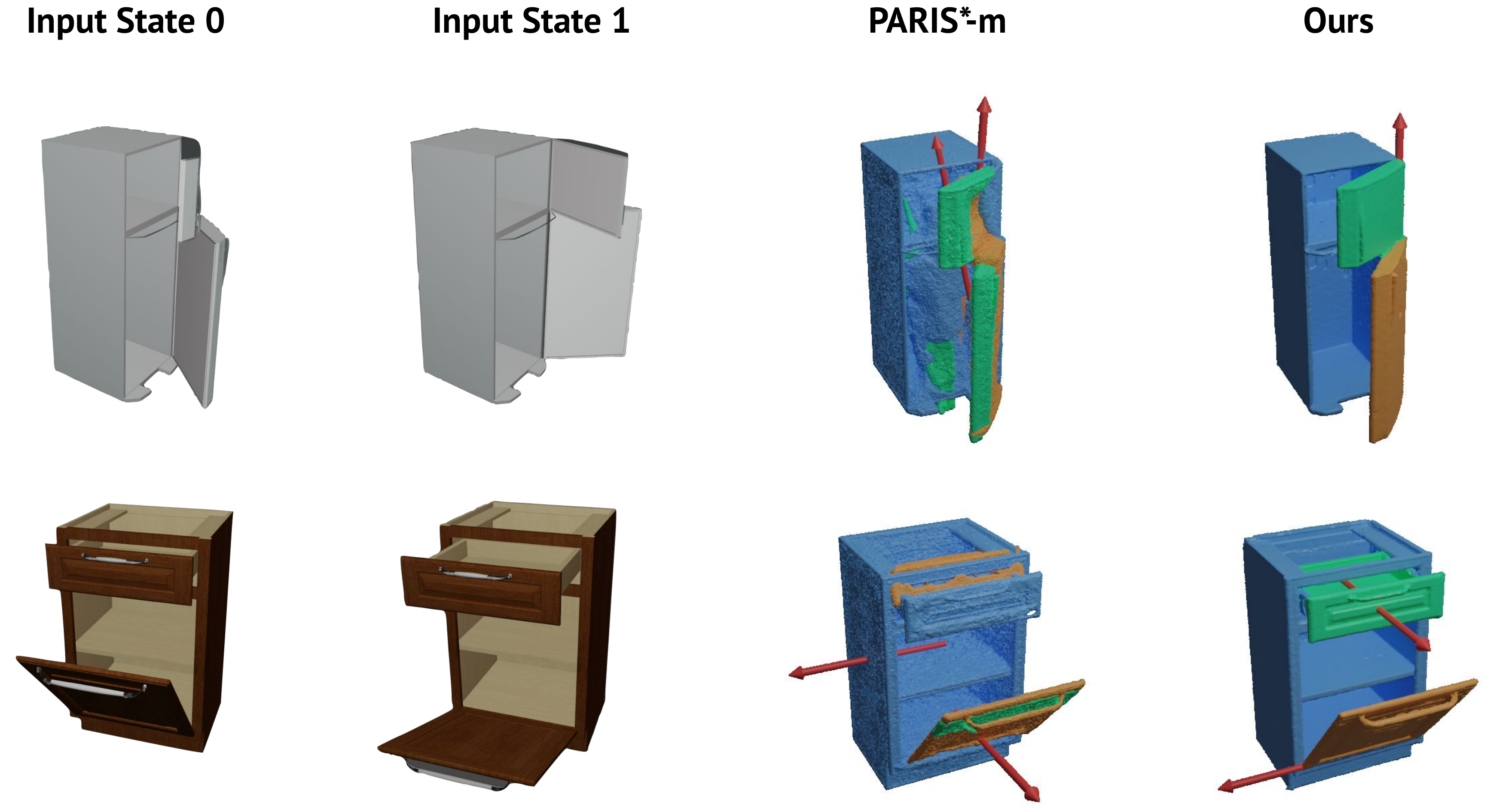}
  \vspace{-5pt}
  \caption{Qualitative results of shape reconstruction,  part segmentation, and joint prediction on multi-part object dataset.}
  \label{fig:qual_mpart}
  \vspace{-20pt}
\end{figure}

\subsection{Datasets}

\boldparagraphstart{PARIS Two-Part Object Dataset.} PARIS \cite{liu2023paris} created a dataset of daily-life two-part articulated objects, including 10 synthetic object instances from PartNet-Mobility \cite{xiang2020sapien} and 2 real-world objects captured with MultiScan~\cite{mao2022multiscan} pipeline. Each object is observed at two joint states, where only one part (``movable part") moves across states, and the other part (``static part") remains static. Observations at each state consist of RGB images and object masks captured from $100$ random views in the upper hemisphere. We additionally rendered depth images for synthetic objects using the same camera parameters as PARIS, and retrieved depth data for real objects from raw RGB-D catpures. 

\boldparagraphstart{Synthetic Multi-Part Object Dataset.} We created 2 synthetic scenes using multi-part instances from PartNet-Mobility \cite{xiang2020sapien}. These objects consist of one static part and multiple movable parts. We capture them at two articulation states, where the multiple movable parts change their individual poses simultaneously across the two states. For each state, we randomly selected 100 views from the upper hemisphere and rendered RGB, depth, and object masks.

\sisetup{detect-all=true}
\begin{table*}[t]
\centering
\def\mywidth{0.99\textwidth} 
\def\myspc{}{\hspace{8ex}}
\resizebox{\mywidth}{!}{
\sisetup{table-auto-round,table-format=.2,table-column-width=1.35cm}
\begin{tabular}{cl|SSSSSSSSSSS[table-column-width=1.6cm]|SSS}

\hline
                               &                                & \multicolumn{11}{c|}{Simulation}                                                                                     & \multicolumn{3}{c}{Real } \bigstrut[t]\\
                               &                                & \multicolumn{1}{c}{FoldChair}                      & \multicolumn{1}{c}{Fridge}                         & \multicolumn{1}{c}{Laptop\textsuperscript{\dag}}   & \multicolumn{1}{c}{Oven\textsuperscript{\dag}}     & \multicolumn{1}{c}{Scissor}                        & \multicolumn{1}{c}{Stapler}                        & \multicolumn{1}{c}{USB}                            & \multicolumn{1}{c}{Washer}                         & \multicolumn{1}{c}{Blade}                          & \multicolumn{1}{c}{Storage\textsuperscript{\dag}}  & \multicolumn{1}{c|}{All}                            & \multicolumn{1}{c}{Fridge}                         & \multicolumn{1}{c}{Storage}                        & \multicolumn{1}{c}{All \bigstrut[b]}\\
\hline

\multirow{4}[2]{*}{\shortstack{Axis\\Ang}} 
& Ditto~\cite{jiang2022ditto}    & 89.35                          & 89.30*                         & 3.12                           & 0.96                           & 4.50                           & 89.86                          & 89.77                          & 89.51                          & 79.54*                         & 6.32                           & 54.22                          & 1.71                           & \bfseries 5.88                  & \bfseries 3.80 \bigstrut[t]\\

& PARIS~\cite{liu2023paris}      & 8.08\tsb{13.2}                 & 9.15\tsb{28.3}                 & \bfseries 0.02\tsb{0.0}         & \bfseries 0.04\tsb{0.0}         & 3.82\tsb{3.4}                  & 39.73\tsb{35.1}                & 0.13\tsb{0.2}                  & 25.36\tsb{30.3}                & 15.38\tsb{14.9}                & \bfseries 0.03\tsb{0.0}         & 10.17\tsb{12.5}                & \bfseries 1.64\tsb{0.3}         & 43.13\tsb{23.4}                & 22.39\tsb{11.9} \\
                               
& PARIS*~\cite{liu2023paris}     & 15.79\tsb{29.3}                & 2.93\tsb{5.3}                  & 0.03\tsb{0.0}                  & 7.43\tsb{23.4}                 & 16.62\tsb{32.1}                & 8.17\tsb{15.3}                 & 0.71\tsb{0.8}                  & 18.40\tsb{23.3}                & 41.28\tsb{31.4}                & \bfseries 0.03\tsb{0.0}         & 11.14\tsb{16.1}                & 1.90\tsb{0.0}                  & 30.10\tsb{10.4}                & 16.00\tsb{5.2} \\
                                
& CSG-reg                        & 0.10\tsb{0.0}                  & 0.27\tsb{0.0}                 & 0.47\tsb{0.0}                 & 0.35\tsb{0.1}                 & 0.28\tsb{0.0}                 & 0.30\tsb{0.0}                 & 11.78\tsb{10.5}               & 71.93\tsb{6.3}                 & 7.64\tsb{5.0}                 & 2.82\tsb{2.5}                 & 9.60\tsb{2.4}                 & 8.92\tsb{0.9}                 & 69.71\tsb{9.6}                & 39.31\tsb{5.2} \\

& 3Dseg-reg                      &\multicolumn{1}{c}{-}                             &\multicolumn{1}{c}{-}                             & 2.34\tsb{0.11}                 &\multicolumn{1}{c}{-}                             &\multicolumn{1}{c}{-}                             &\multicolumn{1}{c}{-}                             &\multicolumn{1}{c}{-}                             &\multicolumn{1}{c}{-}                             & 9.40\tsb{7.5}                 &\multicolumn{1}{c}{-}                             &\multicolumn{1}{c}{-}                             &\multicolumn{1}{c}{-}                             &\multicolumn{1}{c}{-}                             &\multicolumn{1}{c}{-}\\
\rowcolor[rgb]{ .988,  .894,  .839}  \cellcolor[rgb]{1, 1, 1}                               
& Ours & \bfseries    0.03\tsb{0.0}& \bfseries 0.07\tsb{0.0} & 0.06\tsb{0.0} & 0.22\tsb{0.0} & \bfseries 0.11\tsb{0.0} & \bfseries 0.06\tsb{0.0} & \bfseries 0.11\tsb{0.0} & \bfseries 0.43\tsb{0.0} & \bfseries 0.27\tsb{0.0} & 0.06\tsb{0.0} & \bfseries 0.14\tsb{0.0} & 2.10\tsb{0.0} & 18.11\tsb{0.2} & 10.11\tsb{0.1} \bigstrut[b]\\
\hline

\multirow{4}[2]{*}{\shortstack{Axis\\Pos}} & Ditto~\cite{jiang2022ditto}    & 3.77                           & 1.02*                          & 0.01                           & 0.13                           & 5.70                           & 0.20                           & 5.41                           & 0.66                           &\multicolumn{1}{c}{-}                             &\multicolumn{1}{c}{-}                             & 2.11                           & 1.84                           &\multicolumn{1}{c}{-}                             & 1.84 \bigstrut[t]\\

& PARIS~\cite{liu2023paris}      & 0.45\tsb{0.9}                  & 0.38\tsb{1.0}                  & \bfseries 0.00\tsb{0.0}         & \bfseries 0.00\tsb{0.0}         & 2.10\tsb{1.4}                  & 2.27\tsb{3.4}                  & 2.36\tsb{3.4}                  & 1.50\tsb{1.3}                  &\multicolumn{1}{c}{-}                             &\multicolumn{1}{c}{-}                             & 1.13\tsb{1.1}                  & \bfseries 0.34\tsb{0.2}         &\multicolumn{1}{c}{-}                             & \bfseries 0.34\tsb{0.2} \\
                               
& PARIS*~\cite{liu2023paris}     & 0.25\tsb{0.5}                  & 1.13\tsb{2.6}                  & \bfseries 0.00\tsb{0.0}         & 0.05\tsb{0.2}                  & 1.59\tsb{1.7}                  & 4.67\tsb{3.9}                  & 3.35\tsb{3.1}                  & 3.28\tsb{3.1}                  &\multicolumn{1}{c}{-}                             &\multicolumn{1}{c}{-}                             & 1.79\tsb{1.5}                  & 0.50\tsb{0.0}                  &\multicolumn{1}{c}{-}                             & 0.50\tsb{0.0} \\
                                
& CSG-reg                        & 0.02\tsb{0.0}                  & \bfseries 0.00\tsb{0.0}         & 0.20\tsb{0.2}                 & 0.18\tsb{0.0}                 & \bfseries 0.01\tsb{0.0}        & 0.02\tsb{0.0}                  & 0.01\tsb{0.0}                 & 2.13\tsb{1.5}                 &\multicolumn{1}{c}{-}                             &\multicolumn{1}{c}{-}                             & 0.32\tsb{0.2}                 & 1.46\tsb{1.1}                 &\multicolumn{1}{c}{-}                             & 1.46\tsb{1.1} \\
                               
& 3Dseg-reg                      &\multicolumn{1}{c}{-}                             &\multicolumn{1}{c}{-}                             & 0.10\tsb{0.0}                 &\multicolumn{1}{c}{-}                             &\multicolumn{1}{c}{-}                             &\multicolumn{1}{c}{-}                             &\multicolumn{1}{c}{-}                             &\multicolumn{1}{c}{-}                             &\multicolumn{1}{c}{-}                             &\multicolumn{1}{c}{-}                             &\multicolumn{1}{c}{-}                             &\multicolumn{1}{c}{-}                             &\multicolumn{1}{c}{-}                             &\multicolumn{1}{c}{-}\\
                               
\rowcolor[rgb]{ .988,  .894,  .839} \cellcolor[rgb]{1, 1, 1} 
& Ours & \bfseries 0.01\tsb{0.0} & 0.01\tsb{0.0} & \bfseries 0.00\tsb{0.0} & 0.01\tsb{0.0} & 0.02\tsb{0.0} & \bfseries 0.01\tsb{0.0} & \bfseries 0.00\tsb{0.0} & \bfseries 0.01\tsb{0.0} &  \multicolumn{1}{c}{-} & \multicolumn{1}{c}{-} & \bfseries 0.01\tsb{0.0} & 0.57\tsb{0.0} & \multicolumn{1}{c}{-} & 0.57\tsb{0.0} \bigstrut[b]\\
\hline

\multirow{4}[2]{*}{\shortstack{Part\\Motion}} 
& Ditto~\cite{jiang2022ditto}    & 99.36                          & F                              & 5.18                           & 2.09                           & 19.28                          & 56.61                          & 80.60                          & 55.72                          & F                              & 0.09                           & 39.87                          & 8.43                           & 0.38                           & 4.41 \bigstrut[t]\\
                               
& PARIS~\cite{liu2023paris}      & 131.66\tsb{78.9}               & 24.58\tsb{57.7}                & \bfseries 0.03\tsb{0.0}         & \bfseries 0.03\tsb{0.0}         & 120.70\tsb{50.1}               & 110.80\tsb{47.1}               & 64.85\tsb{84.3}                & 60.35\tsb{23.3}                & 0.34\tsb{0.2}                  & 0.30\tsb{0.0}                  & 51.36\tsb{34.2}                & 2.16\tsb{1.1}                  & 0.56\tsb{0.4}                  & 1.36\tsb{0.7} \\
                               
& PARIS*~\cite{liu2023paris}     & 127.34\tsb{75.0}               & 45.26\tsb{58.5}                & \bfseries 0.03\tsb{0.0}         & 9.13\tsb{28.8}                 & 68.36\tsb{64.8}                & 107.76\tsb{68.1}               & 96.93\tsb{67.8}                & 49.77\tsb{26.5}                & 0.36\tsb{0.2}                  & 0.30\tsb{0.0}                  & 50.52\tsb{39.0}                & \bfseries 1.58\tsb{0.0}         & 0.57\tsb{0.1}                  & 1.07\tsb{0.1} \\
                               
& CSG-reg                        & \bfseries 0.13\tsb{0.0}         & 0.29\tsb{0.0}                 & 0.35\tsb{0.0}                 & 0.58\tsb{0.0}                 & 0.20\tsb{0.0}                 & 0.44\tsb{0.0}                 & 10.48\tsb{9.3}                & 158.99\tsb{8.8}               & 0.05\tsb{0.0}                 & 0.04\tsb{0.0}                 & 17.16\tsb{1.8}                & 14.82\tsb{0.1}                & 0.64\tsb{0.1}                 & 7.73\tsb{0.1} \\
                               
& 3Dseg-reg                      &\multicolumn{1}{c}{-}                             &\multicolumn{1}{c}{-}                             & 1.61\tsb{0.1}                 &\multicolumn{1}{c}{-}                             &\multicolumn{1}{c}{-}                             &\multicolumn{1}{c}{-}                             &\multicolumn{1}{c}{-}                             &\multicolumn{1}{c}{-}                             & 0.15\tsb{0.0}                 &\multicolumn{1}{c}{-}                             &\multicolumn{1}{c}{-}                             &\multicolumn{1}{c}{-}                             &\multicolumn{1}{c}{-}                             &\multicolumn{1}{c}{-}\\

\rowcolor[rgb]{ .988,  .894,  .839}  \cellcolor[rgb]{1, 1, 1}                               
& Ours & 0.16\tsb{0.0} & \bfseries 0.09\tsb{0.0} & 0.08\tsb{0.0} & 0.11\tsb{0.0} & \bfseries 0.15\tsb{0.0} & \bfseries 0.05\tsb{0.0} & \bfseries 0.11\tsb{0.0} & \bfseries 0.25\tsb{0.0} & \bfseries 0.00\tsb{0.0} & \bfseries 0.00\tsb{0.0} & \bfseries 0.10\tsb{0.0} & 1.86\tsb{0.0} & \bfseries 0.20\tsb{0.0} & \bfseries 1.03\tsb{0.0} \bigstrut[b]\\
\hline
\multirow{4}[2]{*}{CD-s}       
& Ditto~\cite{jiang2022ditto}    & 33.79                          & 3.05                           & 0.25                           & \bfseries 2.52                  & 39.07                          & 41.64                          & 2.64                           & 10.32                          & 46.9                           & 9.18                           & 18.94                          & 47.01                          & 16.09                          & 31.55 \bigstrut[t]\\
                               
& PARIS~\cite{liu2023paris}      & 9.16\tsb{5.0}                  & 3.65\tsb{2.7}                  & \bfseries 0.16\tsb{0.0}         & 12.95\tsb{1.0}                 & 1.94\tsb{3.8}                  & \bfseries 1.88\tsb{0.2}         & 2.69\tsb{0.3}                  & 25.39\tsb{2.2}                 & 1.19\tsb{0.6}                  & 12.76\tsb{2.5}                 & 7.18\tsb{1.8}                  & 42.57\tsb{34.1}                & 54.54\tsb{30.1}                & 48.56\tsb{32.1} \\
                               
& PARIS*~\cite{liu2023paris}     & 10.20\tsb{5.8}                 & 8.82\tsb{12.0}                 & \bfseries 0.16\tsb{0.0}         & 3.18\tsb{0.3}                  & 15.58\tsb{13.3}                & 2.48\tsb{1.9}                  & \bfseries 1.95\tsb{0.5}         & 12.19\tsb{3.7}                 & 1.40\tsb{0.7}                  & 8.67\tsb{0.8}                  & 6.46\tsb{3.9}                  & 11.64\tsb{1.5}                 & 20.25\tsb{2.8}                 & 15.94\tsb{2.1} \\
                               
& CSG-reg                        & 1.69                  & 1.45                  & 0.32         & 3.93         & 3.26                  & 2.22         & \bfseries 1.95         & \bfseries 4.53         & 0.59                  & 7.06                  & 2.70                  & 6.33                  & 12.55                 & 9.44 \\
                               
& 3Dseg-reg                      &\multicolumn{1}{c}{-}                             &\multicolumn{1}{c}{-}                             & 0.76                  &\multicolumn{1}{c}{-}                             &\multicolumn{1}{c}{-}                             &\multicolumn{1}{c}{-}                             &\multicolumn{1}{c}{-}                             &\multicolumn{1}{c}{-}                             & 66.31                 &\multicolumn{1}{c}{-}                             &\multicolumn{1}{c}{-}                             &\multicolumn{1}{c}{-}                             &\multicolumn{1}{c}{-}                             &\multicolumn{1}{c}{-}\\

\rowcolor[rgb]{ .988,  .894,  .839}  \cellcolor[rgb]{1, 1, 1}                               
& Ours & \bfseries 0.18\tsb{0.0} & \bfseries 0.60\tsb{0.0} & 0.32\tsb{0.0} & 4.66\tsb{0.0} & \bfseries 0.40\tsb{0.1} & 2.65\tsb{0.0} & 2.19\tsb{0.0} & 4.80\tsb{0.0} & \bfseries 0.55\tsb{0.0} & \bfseries 4.69\tsb{0.0} & \bfseries 2.10\tsb{0.0} & \bfseries 2.53\tsb{0.0} & \bfseries 10.86\tsb{0.1} & \bfseries 6.69\tsb{0.0} \bigstrut[b]\\
\hline
\multirow{4}[2]{*}{CD-m}       
& Ditto~\cite{jiang2022ditto}    & 141.11                         & 0.99                           & 0.19                           & 0.94                           & 20.68                          & 31.21                          & 15.88                          & 12.89                          & 195.93                         & 2.20                           & 42.20                          & 50.60                          & \bfseries 20.35                 & 35.48 \bigstrut[t]\\
                               
& PARIS~\cite{liu2023paris}      & 8.99\tsb{7.6}                  & 7.76\tsb{11.2}                 & 0.21\tsb{0.2}                  & 28.70\tsb{15.2}                & 46.64\tsb{40.7}                & 19.27\tsb{30.7}                & 5.32\tsb{5.9}                  & 178.43\tsb{131.7}              & 25.21\tsb{9.5}                 & 76.69\tsb{6.1}                 & 39.72\tsb{25.9}                & 45.66\tsb{31.7}                & 864.82\tsb{382.9}              & 455.24\tsb{207.3} \\
                               
& PARIS*~\cite{liu2023paris}     & 17.97\tsb{24.9}                & 7.23\tsb{11.5}                 & \bfseries 0.15\tsb{0.0}         & 6.54\tsb{10.6}                 & 16.65\tsb{16.6}                & 30.46\tsb{37.0}                & 10.17\tsb{6.9}                 & 265.27\tsb{248.7}              & 117.99\tsb{213.0}              & 52.34\tsb{11.0}                & 52.48\tsb{58.0}                & 77.85\tsb{26.8}                & 474.57\tsb{227.2}              & 276.21\tsb{127.0} \\
                                
& CSG-reg                        & 1.91                  & 21.71                 & 0.42                  & 256.99                & 1.95                  & 6.36                  & 29.78                 & 436.42                & 26.62                 & 1.39                  & 78.36                 & 442.17                & 521.49                & 481.83 \\
                               
& 3Dseg-reg                      &\multicolumn{1}{c}{-}                             &\multicolumn{1}{c}{-}                             & 1.01                  &\multicolumn{1}{c}{-}                             &\multicolumn{1}{c}{-}                             &\multicolumn{1}{c}{-}                             &\multicolumn{1}{c}{-}                             &\multicolumn{1}{c}{-}                             & 6.23                  &\multicolumn{1}{c}{-}                             &\multicolumn{1}{c}{-}                             &\multicolumn{1}{c}{-}                             &\multicolumn{1}{c}{-}                             &\multicolumn{1}{c}{-}\\
\rowcolor[rgb]{ .988,  .894,  .839}  \cellcolor[rgb]{1, 1, 1}                               
& Ours & \bfseries 0.15\tsb{0.0} & \bfseries 0.27\tsb{0.0} & 0.16\tsb{0.0} & \bfseries 0.47\tsb{0.0} & \bfseries 0.41\tsb{0.0} & \bfseries 2.27\tsb{0.0} & \bfseries 1.34\tsb{0.0} & \bfseries 0.36\tsb{0.0} & \bfseries 1.50\tsb{0.1} & \bfseries 0.37\tsb{0.0} & \bfseries 0.73\tsb{0.0} & \bfseries 1.14\tsb{0.0} & 26.46\tsb{5.0} & \bfseries 13.80\tsb{2.5} \bigstrut[b]\\
\hline
\multirow{4}[2]{*}{CD-w}       
& Ditto~\cite{jiang2022ditto}    & 6.80                           & 2.16                           & 0.31                           & \bfseries 2.51                  & 1.70                           & 2.38                           & 2.09                           & 7.29                           & 42.04                          & \bfseries 3.91                  & 7.12                           & 6.50                           & 14.08                          & 10.29 \bigstrut[t]\\
                               
& PARIS~\cite{liu2023paris}      & 1.80\tsb{1.2}                  & 2.92\tsb{0.9}                  & 0.30\tsb{0.1}                  & 11.73\tsb{1.1}                 & 10.49\tsb{20.7}                & 3.58\tsb{4.2}                  & 2.00\tsb{0.2}                  & 24.38\tsb{3.3}                 & 0.60\tsb{0.2}                  & 8.57\tsb{0.4}                  & 6.64\tsb{3.2}                  & 22.98\tsb{15.5}                & 63.35\tsb{22.2}                & 43.16\tsb{18.9} \\
                               
& PARIS*~\cite{liu2023paris}     & 4.37\tsb{6.4}                  & 5.53\tsb{4.7}                  & \bfseries 0.26\tsb{0.0}         & 3.18\tsb{0.3}                  & 3.90\tsb{3.6}                  & 5.27\tsb{5.9}                  & 1.78\tsb{0.2}                  & 10.11\tsb{2.8}                 & 0.58\tsb{0.1}                  & 7.80\tsb{0.4}                  & 4.28\tsb{2.4}                  & 8.99\tsb{1.4}                  & 32.10\tsb{8.2}                 & 20.55\tsb{4.8} \\
                                
& CSG-reg                        & 0.48                  & 0.98                  & 0.40                  & 3.00         & 1.70                  & \bfseries 1.99         & 1.20                  & \bfseries 4.48         & 0.56                  & 4.00                  & 1.88                  & 5.71                  & 14.29                 & 10.00 \\
                               
& 3Dseg-reg                      &\multicolumn{1}{c}{-}                             &\multicolumn{1}{c}{-}                             & 0.81                  &\multicolumn{1}{c}{-}                             &\multicolumn{1}{c}{-}                             &\multicolumn{1}{c}{-}                             &\multicolumn{1}{c}{-}                             &\multicolumn{1}{c}{-}                             & 0.78                  &\multicolumn{1}{c}{-}                             &\multicolumn{1}{c}{-}                             &\multicolumn{1}{c}{-}                             &\multicolumn{1}{c}{-}                             &\multicolumn{1}{c}{-}\\
\rowcolor[rgb]{ .988,  .894,  .839}  \cellcolor[rgb]{1, 1, 1}                               
& Ours & \bfseries 0.27\tsb{0.0} & \bfseries 0.70\tsb{0.0} & 0.35\tsb{0.0} & 4.18\tsb{0.0} & \bfseries 0.43\tsb{0.0} & 2.19\tsb{0.0} & \bfseries 1.18\tsb{0.0} & 4.74\tsb{0.0} & \bfseries 0.36\tsb{0.0} & 3.99\tsb{0.0} & \bfseries 1.84\tsb{0.0} & \bfseries 2.19\tsb{0.0} & \bfseries 9.33\tsb{0.0} & \bfseries 5.76\tsb{0.0} \bigstrut[b]\\
\hline
\end{tabular}%
}
\vspace{-0.1in}
\caption{Results on PARIS dataset including both synthetic and real data.  
(Shown are the average $\pm$ stdev over 10 trials with different random seeds; see supplementary for details.)
PARIS*~\cite{liu2023paris} is augmented with depth for fair comparison under the same input modality. Objects with \dag \ are the seen categories that Ditto~\cite{hsu2023ditto} has been trained on.   Ditto sometimes gives wrong motion type predictions, which are noted with F for joint state and * for joint axis or position.  Note that Blade, Storage, and Real Storage have prismatic joints, so there is no Axis Pos.
}
\label{tab:two_part}
\end{table*}
\begin{table*}[t]
\centering
\def\mywidth{0.95\textwidth} 
\resizebox{\mywidth}{!}{

\begin{tabular}{lc|rrrrrrrrrr}
\hline
                               &                                & \multicolumn{1}{c}{Axis Ang 0} & \multicolumn{1}{c}{Axis Ang 1} & \multicolumn{1}{c}{Axis Pos 0} & \multicolumn{1}{c}{Axis Pos 1} & \multicolumn{1}{c}{Part Motion 0} & \multicolumn{1}{c}{Part Motion 1} & \multicolumn{1}{c}{CD-s}       & \multicolumn{1}{c}{CD-m 0}     & \multicolumn{1}{c}{CD-m 1}     & \multicolumn{1}{c}{CD-w} \bigstrut\\
\hline
\multirow{2}[2]{*}{Fridge-m} & PARIS*-m~\cite{liu2023paris}     & 34.52                          & 15.91                          & 3.60                           & 1.63                           & 86.21                          & 105.86                         & 8.52                           & 526.19                         & 160.86                         & 15.00 \bigstrut[t]\\
                               & \cellcolor[rgb]{ .988,  .894,  .839}Ours & \cellcolor[rgb]{ .988,  .894,  .839}\textbf{0.16} & \cellcolor[rgb]{ .988,  .894,  .839}\textbf{0.10} & \cellcolor[rgb]{ .988,  .894,  .839}\textbf{0.01} & \cellcolor[rgb]{ .988,  .894,  .839}\textbf{0.00} & \cellcolor[rgb]{ .988,  .894,  .839}\textbf{0.11} & \cellcolor[rgb]{ .988,  .894,  .839}\textbf{0.13} & \cellcolor[rgb]{ .988,  .894,  .839}\textbf{0.61} & \cellcolor[rgb]{ .988,  .894,  .839}\textbf{0.40} & \cellcolor[rgb]{ .988,  .894,  .839}\textbf{0.52} & \cellcolor[rgb]{ .988,  .894,  .839}\textbf{0.89} \bigstrut[b]\\
\hline
\multirow{2}[2]{*}{Storage-m} & PARIS*-m~\cite{liu2023paris}     & 43.26                          & 26.18                          & 10.42                          & -                              & 79.84                          & 0.64                           & 8.56                           & 128.62                         & 266.71                         & 8.66 \bigstrut[t]\\
                               & \cellcolor[rgb]{ .988,  .894,  .839}Ours & \cellcolor[rgb]{ .988,  .894,  .839}\textbf{0.21} & \cellcolor[rgb]{ .988,  .894,  .839}\textbf{0.88} & \cellcolor[rgb]{ .988,  .894,  .839}\textbf{0.05} & \cellcolor[rgb]{ .988,  .894,  .839}\textbf{-} & \cellcolor[rgb]{ .988,  .894,  .839}\textbf{0.13} & \cellcolor[rgb]{ .988,  .894,  .839}\textbf{0.00} & \cellcolor[rgb]{ .988,  .894,  .839}\textbf{0.85} & \cellcolor[rgb]{ .988,  .894,  .839}\textbf{0.21} & \cellcolor[rgb]{ .988,  .894,  .839}\textbf{3.46} & \cellcolor[rgb]{ .988,  .894,  .839}\textbf{0.99} \bigstrut[b]\\
\hline
\end{tabular}%

}
\vspace{-0.1in}
\caption{Results on multi-part object dataset, averaged over 10 trials with different random seeds; see supplementary for details.
PARIS*-m~\cite{liu2023paris} is augmented with depth and extended to handle objects with more than two parts. Joint 1 of ``Storage-m'' is prismatic, so there is no Axis Pos.
}
\label{tab:multi_part}
\vspace{-15pt}
\end{table*}

\begin{table}[t]
\centering
\def\mywidth{0.45\textwidth} 
\resizebox{\mywidth}{!}{
\sisetup{table-auto-round,table-format=.2,table-column-width=1.5cm}
\begin{tabular}{l|SSSSS}
\hline
                               & \multicolumn{1}{c}{Axis Ang}                       & \multicolumn{1}{c}{Axis Pos}                       & \multicolumn{1}{c}{Part Motion}                    & \multicolumn{1}{c}{CD-s}                           & \multicolumn{1}{c}{CD-m} \bigstrut\\
\hline
Proposed                       & 0.26\tsb{0.04}                 & 0.01\tsb{0.00}                 & 0.09\tsb{0.01}                 & 0.80\tsb{0.08}                  & 0.33\tsb{0.00} \bigstrut[t]\\
w/o sharem                     & 14.42\tsb{0.90}                & 0.01\tsb{0.00}                 & 0.28\tsb{0.02}                 & 2.42\tsb{0.06}                 & 127.08\tsb{17.38} \\
w/o matching                   & 4.34\tsb{0.08}                 & 1.14\tsb{0.13}                 & 8.06\tsb{0.06}                 & 1.05\tsb{0.02}                 & 112.12\tsb{4.39} \\
w/o rgb                        & 0.43\tsb{0.11}                 & 0.01\tsb{0.00}                 & 0.13\tsb{0.01}                 & 0.76\tsb{0.03}                 & 0.35\tsb{0.00} \\
w/o occ                        & 0.42\tsb{0.14}                 & 0.01\tsb{0.00}                 & 0.12\tsb{0.02}                 & 7.16\tsb{5.73}                 & 164.33\tsb{92.62} \\
w/o collision                  & 0.18\tsb{0.03}                 & 0.01\tsb{0.00}                 & 0.10\tsb{0.02}                  & 11.64\tsb{0.67}                & 191.76\tsb{2.08} \bigstrut[b]\\
\hline
\end{tabular}%
}
\vspace{-0.1in}
\caption{Ablation study on multi-part objects. We report results averaged over all joints/movable parts across 5 trials. \textit{w/o shared motion} optimizes two separate sets of motion parameters for the two motion models at state $0$ and $1$.
Other ablated versions omit the matching, RGB consistency, occupancy consistency, and collision loss, respectively.}
\label{tab:ablation}
\vspace{-20pt}
\end{table}

\subsection{Metrics}

\boldparagraphstart{Object- and Part-level Geometry.} We evaluate object and part mesh reconstructions with bi-directional Chamfer-$\mathit{l}_1$ distance (CD),  by sampling 10K points uniformly on the groundtruth and predicted meshes. We report \textbf{CD-w (mm)} for the whole object, \textbf{CD-s (mm)} for the static part, and \textbf{CD-m (mm)} for the movable part. Following \cite{liu2023paris}, we report these values in millimeters. 

\boldparagraphstart{Articulation Model and Cross-State Part Motion.} We evaluate the estimated articulation model with \textbf{Axis Ang Err ($^\circ$)}, the angular error of the predicted joint axis for both revolute and prismatic joints, and \textbf{Axis Pos Err (0.1m)}, the minimum distance between the predicted and ground-truth joint axes for revolute joints. We also evaluate the estimated part motion between states with \textbf{Part Motion Err ($^\circ$ or m)} (referred to as Joint State by \cite{liu2023paris}), the geodesic distance error of predicted rotations for revolute joints, or the Euclidean distance error of translations for prismatic joints.

\subsection{Baselines}

\boldparagraphstart{Ditto} \cite{jiang2022ditto} is a feed-forward model that reconstructs part-level meshes and the motion model (joint type, axis, and state) of a two-part articulated object given multi-view fused point cloud observations at two different joint states. It shares the same assumption as  PARIS \cite{liu2023paris} that only one object part moves across states. We follow \cite{liu2023paris}'s protocol and report results from Ditto's released model pretrained on 4 object categories from Shape2Motion \cite{wang2019shape2motion}. 

\boldparagraphstart{PARIS}~\cite{liu2023paris} reconstructs part-level shape and appearance as well as the motion model of a two-part articulated object, given multi-view RGB observations at two articulation states. It adopts a NeRF-based representation and performs per-object optimization such that it can be applied to arbitrary unknown objects. The object is modeled as the composition of a static part field and a mobile part field, as well as a transformation of the mobile part field that explains cross-state changes. Part fields and transformation are jointly optimized with an image rendering loss.

\boldparagraphstart{PARIS*.} For fair comparison to our RGBD-based approach, we augmented PARIS with depth supervision following \cite{deng2022depthsupervised} and denote this version PARIS*.

\boldparagraphstart{PARIS*-m.} 
The original PARIS~\cite{liu2023paris} is limited to two-part objects.
To make it applicable to more general $P$-part objects, we modified PARIS to optimize a static field and $P - 1$ mobile fields, as well as their $P - 1$ cross-state rigid transformations. We also augmented it with depth supervision.

\boldparagraphstart{CSG-reg.} It reconstructs the object at each state with TSDF fusion, and applies Constructive Solid Geometry to per-state TSDF to get static (intersection) and movable (difference) parts similar to Ours-ICP baseline in PARIS. It then performs Fast Global Registration~\cite{zhou2016fast} and colored ICP~\cite{park2017colored} to align movable parts and estimate joint motions. 

\boldparagraphstart{3Dseg-reg.} It follows the same procedure as CSG-reg, but uses pretrained 3D object part segmentation model PAConv~\cite{xu2021paconv} to segment the reconstructed objects. Since PAConv does not generalize well to unseen categories, we only report numbers for trained categories laptop and blade. 

\subsection{Experiment and Evaluation Setup.} 

We follow \cite{jiang2022ditto,liu2023paris}'s setting where part $0$ is assumed to remain static ($R_0 = I, \textbf{t}_0 = \textbf{0}$). For multi-part objects, the number of parts is assumed known to all evaluated methods but joint types are unknown. In order to find the corresponding part for evaluation, we iterate through all possible pairs between predicted and ground-truth parts and report the best match with the smallest total chamfer distance. To remove floaters that disproportionately affect the chamfer distance, we apply a mesh clustering post-processing step to all methods, where we remove connected mesh components with less than $\tau=10\%$ of the vertices of the largest cluster. Following \cite{liu2023paris}, we transform our extracted parts with predicted motions to state $t=0$ for evaluation.

We observed that optimization-based methods such as PARIS may yield different final results with different random initializations of the model. For comprehensive evaluation, we run all optimization-based methods $10$ times with different random seeds, reporting the mean and standard deviation for each metric across the $10$ trials. Please refer to the appendix for more statistics. 

\subsection{Results on PARIS Two-Part Object Dataset}\label{sec:result_two}
 
Table \ref{tab:two_part} shows results on PARIS Two-Part Object Dataset including synthetic and real instances, summarized over 10 trials. 
Ditto relies on object shape and structure priors learned from training categories, which cover laptop, oven, and storage in evaluation. While it does well on seen categories, especially on shape reconstructions, a clear generalization gap can be observed in terms of unseen categories. 
PARIS exhibits large performance variances across trials for most instances. While it performs good reconstruction in some trials, it occasionally fails drastically, leading to overall much worse performance on both shape and articulation reconstruction. The depth supervision in PARIS* improves object-level shape reconstruction, bringing significant improvement in CD-w on challenging objects such as oven, scissor, washer from synthetic data and all  real instances. 
At the same time, depth further complicates the optimization, leading to more failure cases and larger variance, resulting in worse average articulation predictions. Both CSG-reg and 3Dseg-reg do well on easy objects such as synthetic laptops but struggle elsewhere. Notably, segmentation errors (e.g., intersections containing the movable part of the blade, noise mistaken as movable part) easily propagate into traditional registration-based articulation estimation. 

Our method is robust to initializations and consistently achieves accurate shape and articulation reconstruction across trials. We perform better than baseline methods on most instances. 
As shown in Fig.~\ref{fig:paris_syn}, our approach accurately reconstructs the part geometry and joint axis of real and synthetic objects, while baselines suffer from segmentation noises or complete failures.

\subsection{Results on Multi-Part Objects}

Table \ref{tab:multi_part} summarizes results on multi-part objects across 10 trials. As shown in Fig. \ref{fig:qual_mpart}, with the increased complexity in object structure, PARIS*-m fails to correctly segment the object even coarsely and performs poorly on both shape and articulation reconstruction. PARIS' single image rendering objective fails to drive the optimization process to the correct solution. In contrast, our method achieves high-quality reconstruction with the help of rich information from 2D images, 3D geometries, and kinematics.

\subsection{Ablation Study}

We examine the effectiveness of our design choices on multi-part objects since they are more challenging. We report joint prediction and part reconstruction metrics averaged over all joints/movable parts and instances across $5$ random trials. As shown in Table \ref{tab:ablation}, sharing motions between the two articulation models significantly improves all metrics by leveraging information from both directions. Matching loss also effectively helps guide articulation reconstruction. Collision loss and occupancy consistency loss both add constraints to the whole space beyond surface region, improving dense field learning and thus resulting in higher-quality part-level reconstructions. 

\section{Conclusion}

We presented a framework that reconstructs the geometry and articulation model of unknown articulated objects given two scans of the object at different joint states. It is per-object optimized and applies to arbitrary articulated objects without assuming any category or articulation priors. It also handles more than one movable part. 
Our proposed two-stage approach disentangles the problems of object-level shape reconstruction and articulation reasoning. By enforcing a set of carefully designed loss terms on a point correspondence field derived from the articulation model, our method effectively leverages cues from image feature matching, object geometry reconstructions, as well as kinematic rules. Extensive experiments indicate our approach achieves more accurate and stable results than prior work. However, challenges remain in applying the method to more general settings, e.g., where camera poses are unknown and object base parts are unaligned. How to fuse observations at different states when the occlusion pattern changes is also an interesting open problem. We leave them to future work.

\clearpage
{
    \small
    \bibliographystyle{ieeenat_fullname}
    \bibliography{main}
}

\clearpage
\maketitlesupplementary

\appendix

\section{Method Details}

In this section, we describe the details of our method and adapted PARIS*/PARIS*-m.
We will also release our code and data to facilitate future research. 

\subsection{Neural Object Field}\label{sec:supp_nof}

We use Neural Object Field \cite{wen2023bundlesdf} as the object representation for our stage one reconstruction. We follow the practice of \cite{wen2023bundlesdf} and we describe the specifics below. 

Given multi-view posed RGB-D images of the object $\mathcal{O}^t$ at state $t, t\in \{0, 1\}$, we reconstruct the object in the form of a Neural Object Field~\cite{wen2023bundlesdf} $(\Omega^t, \Phi^t)$ (we omit $t$ for simplicity in the following). The geometry network $\Omega: \mathbf{x} \mapsto s$ maps spatial point $\mathbf{x} \in \mathbb{R}^3$ to its truncated signed distance $s = \operatorname{clip}(d / \tau_{d}, -1, 1)$, where $\tau_{d} = 0.03$ is the truncation distance, $d$ is the signed distance to the object surface. The appearance network $\Phi: (\mathbf{x}, \mathbf{d}) \mapsto \mathbf{c}$ maps point $\mathbf{x} \in \mathbb{R}^3$ and view direction $\mathbf{d} \in \mathbb{S}^2$ to RGB color $C \in \mathbb{R}^3$. 

We supervise the Neural Object Field at 3D points $\{\mathbf{x}_i = \mathbf{o} + t_i\mathbf{d}\}$ sampled along camera rays $\mathbf{r}(t) = \mathbf{o} + t\mathbf{d}$, where $\mathbf{o} \in \mathbb{R}^3$ denotes ray origin, and $\mathbf{d} \in \mathbb{S}^2$ denotes ray direction.

The expected color $C(\mathbf{r})$ is approximated by a weighted average of point colors around the object surface:
\begin{align}
&C(\mathbf{r}) = \mathbb{E}_{\mathbf{x}_i \in \mathcal{X}_{\text{surf}}} \left[ w(\mathbf{x}_i) \Phi(\mathbf{x}_i, \mathbf{d})\right], \\
&\mathcal{X}_\text{surf} = \{\mathbf{x} | |\Omega(\mathbf{x})| < 1\}, \\
&w(\mathbf{x}_i) = \frac{1}{(1 + e^{-\alpha\Omega(\mathbf{x}_i)})(1 + e^{\alpha\Omega(\mathbf{x}_i)})},\label{eq:nof_render_weight}
\end{align}

where $\mathcal{X}_\text{surf}$ denotes the set of points within truncation distance $\tau_{d}$ to the object surface, $w(\mathbf{x}_i)$ is a bell-shaped function that peaks at object surface, $\alpha = 5$ is a hyperparameter that controls its sharpness.

Let $z(\mathbf{r})$, $\hat{C}(\mathbf{r})$ be the groundtruth depth and color at training ray $\mathbf{r} \in \mathcal{R}$, $d(\mathbf{x})$ be $\mathbf{x}$'s distance to ray origin $\mathbf{o}$, $\hat\Omega(\mathbf{x})$ be the groundtruth untruncated SDF. We supervise $(\Omega, \Phi)$ with color rendering loss $\mathcal{L}_{\text{render}}$ (denoted $\mathcal{L}_c$ in the main paper, changed to avoid confusion with $l_{c}$ in consistency loss):  
\begin{align}
\mathcal{L}_{\text{render}} =  \mathbb{E}_{\mathbf{r} \in \mathcal{R}}\left[||C(\mathbf{r}) - \hat{C}(\mathbf{r})||_{2}^2\right],
\end{align}

And SDF loss $\mathcal{L}_{\text{SDF}}$:

\begin{align}
&\mathcal{L}_{\text{SDF}} = \lambda_{e} \mathcal{L}_e + \lambda_\text{surf} \mathcal{L}_\text{surf},\\
&\mathcal{L}_e = \mathbb{E}_{\mathbf{x} \in \mathcal{X}_{e}} \left[|\Omega(\mathbf{x}) - 1|\right],\\
&\mathcal{L}_\text{surf} = \mathbb{E}_{\mathbf{x} \in \mathcal{X}_\text{surf}} \left[\left(\Omega(\mathbf{x})\cdot\tau_d - (z(\mathbf{r}) - d(\mathbf{x})))\right)^2\right],
\end{align}
where $\mathcal{X}_{e} = \{\mathbf{x} | \hat\Omega(\mathbf{x}) > \tau_{d} \}$ denotes the empty space in front of the object surface, $\Omega(\mathbf{x})\cdot\tau_{d}$ is the predicted SDF, $\left(z(\mathbf{r}) - d(\mathbf{x})\right)$ approximates groundtruth SDF for points in the near-surface region $\mathcal{X}_\text{surf}$. 
For more stable training, we substitute predicted signed distance $\Omega(\mathbf{x}_i)$ in Eq.~(\ref{eq:nof_render_weight}) with approximated groundtruth signed distance $(z(\mathbf{r}) - d(\mathbf{x}))$. 

The total loss for training Neural Object Field in the first stage is
\begin{align}
\mathcal{L} = \lambda_{\text{render}}\mathcal{L}_{\text{render}} + \lambda_{\text{SDF}}\mathcal{L}_{\text{SDF}}
\end{align}

We set $\lambda_{\text{render}} = 10, ~\lambda_{\text{SDF}} = 1, ~\lambda_{e} = 1~,\lambda_{\text{surf}}= 6000~$, following \cite{wen2023bundlesdf}. We also build an Octree from depth inputs to speed up ray sampling following their practice.

\subsection{Architecture Details}\label{sec:supp_model_details}

Neural Object Field $(\Omega, \Phi)$ is implemented with multi-resolution hash encoding \cite{muller2022instant} of the input position $\mathbf{x}$, spherical embedding of the input view direction $\mathbf{d}$, followed by a 2-layer MLP for TSDF and a 3-layer MLP for color. 

The segmentation field $\mathcal{P}(\mathbf{x}, i)$ is implemented with a dense voxel feature grid followed by a 3-layer MLP. The raw $P$-dim output is activated with softmax to get a probability distribution over $P$ parts. The dense voxel feature grid has size $50 \times 50 \times 50$ and feature dimension $C=20$.

The hidden dimension of all MLPs is set to $64$.

\subsection{Training Details}\label{sec:supp_train_details}

\boldparagraphstart{Generating Mesh, SDF, and Occupancy Field} We extract meshes from the TSDF field $\Omega$ using marching cubes~\cite{lorensen1998marching} with resolution $0.003$. We compute the smoothed occupancy value from SDF with $s = 0.01$ in Eq.~(\ref{eq:occupancy}). For computation efficiency, we pre-computed SDF values at a grid of resolution $0.01$, and use trilinear interpolation of the precomputed values for arbitrary query $\mathbf{x}$.

\boldparagraphstart{Handling Visibility} We set $\epsilon=0.03$ in Eq.~(\ref{eq:visibility}) for visibility computation, and discount invisible corresponding points in the consistency loss with $w_{vis} = 0.5$.

\boldparagraphstart{Training Parameters} We set $\alpha=5$ in Eq.~(\ref{eq:render_weight}) while computing point weights for near-surface points in consistency loss, consistent with the setting of $\alpha$~Eq.~(\ref{eq:nof_render_weight}) in the first stage rendering loss. We set $\lambda_s=10,~\lambda_c=0.1,~\lambda_o=5$ in consistency loss Eq.~(\ref{eq:consistency_loss}), $\lambda_{\text{cns}}=1,~\lambda_{\text{match}}=500,~\lambda_{\text{coll}}=50$ in total training loss Eq.~(\ref{eq:total_loss}). 

We implement our method with PyTorch and use Adam optimizer with an initial learning rate of $0.01$ and exponential decay with factor $0.1$. Each stage of the reconstruction optimizes for $2000$ steps. We also enable occupancy consistency loss and collision loss (both aim at better part segmentation and make more sense when joint parameters are roughly optimized) after $500$ optimization steps in the second stage.

\boldparagraphstart{Computation Time} The optimization part of our method runs for 40 minutes on an NVIDIA Tesla V100 GPU. Precomputing SDF takes another 20 minutes, which we plan to optimize with parallel computation.

\subsection{Inference Details}\label{sec:supp_inference_details}

To extract mesh for part $i$, we run marching cubes on the reconstructed SDF field $\hat{\Omega}(\mathbf{x})$, during which we additionally query the part index of the grid points and set SDF values of points not belonging to part $i$ to $1$ (out of the part). 

Given the raw optimized 6-DoF rigid transformation $(R, \mathbf{t})$, we classify the joint as prismatic when $|\operatorname{angle}(R^0_i)| < \tau_{r}$, where the threshold $\tau_r=10^{\circ}$. For prismatic joints, we take the translation component $\mathbf{t}$ for axis and part motion computation. For revolute joints, we compute axis direction $\mathbf{u}$ and rotation angle $\theta$ from the rotation component $R$, and compute axis position $\mathbf{p}$ as the solution of $\arg\min_{\mathbf{p}} ||(I -R)\mathbf{p} - \mathbf{t}||_{2}^2$.

\subsection{PARIS*/PARIS*-m Implementation Details}\label{sec:supp_paris_details}

We augment original PARIS with depth supervision following the practice of \cite{deng2022depthsupervised}, where the following depth rendering loss is added to the optimization. 
\begin{align}
\mathcal{L}_d &= ||\hat{D}(\mathbf{r}) - D_{gt}||^2_2,\\
\hat{D}(\mathbf{r}) &= \sum_{i = 1}^NT_i(1 - \exp(-\sigma(t_i)\delta_i))t_i,\\
T_i &= \exp(-\sum_{j=1}^{i - 1}\sigma(t_j)\delta_j),\\
\delta_j &= t_{j + 1} - t_j
\end{align}
We use a loss weight of $\lambda_{d} = 0.01$ since larger weights tend to sabotage the optimization and smaller weights have very limited impact.

\begin{table*}[h!]
\centering
\def\mywidth{0.99\textwidth} 
\def\myspc{}{\hspace{8ex}}
\resizebox{\mywidth}{!}{

\begin{tabular}{cl|cccccccccc|cc}
\hline
                              &                                     & \multicolumn{10}{c|}{Simulation}                                                           & \multicolumn{2}{c}{Real} \bigstrut[t]\\
                               &                                & FoldChair                      & Fridge                         & Laptop   & Oven     & Scissor                        & Stapler                        & USB                            & Washer                         & Blade                          & Storage                          & Fridge                         & Storage         \bigstrut[b]\\
\hline
\multirow{3}[2]{*}{\shortstack{Axis\\Ang}}     & PARIS\textsuperscript{\dag}~\cite{liu2023paris} & 0.02      & 0.00   & 0.03   & 0.03  & 0.02    & 0.07    & 0.07 & 0.08   & 0.00  & 0.37    & 1.91       & 3.88        \bigstrut[t]\\
                             & PARIS~\cite{liu2023paris}           & 0.03      & 0.00   & 0.00   & 0.04  & 0.03    & 42.99   & 0.00 & 0.04   & 0.11  & 0.02    & 1.90       & 14.61       \\
                              & PARIS*~\cite{liu2023paris}          & 0.02      & 0.00   & 0.03   & 0.04  & 0.00    & 1.05    & 0.02 & 0.08   & 1.66  & 0.03    & 1.91       & 15.64       
                          \bigstrut[b]\\
\hline
\multirow{3}[2]{*}{\shortstack{Axis\\Pos}}     & PARIS\textsuperscript{\dag}~\cite{liu2023paris} & 0.00      & 0.00   & 0.00   & 0.00  & 0.00    & 0.01    & 0.00 & 0.02   &    -  &    -    & 0.53       &       -    \bigstrut[t]\\
                              & PARIS~\cite{liu2023paris}           & 0.00      & 0.00   & 0.00   & 0.00  & 0.00    & 0.16    & 0.01 & 0.01   & -  & -    & 0.54       & -        \\
                              & PARIS*~\cite{liu2023paris}          & 0.00      & 0.00   & 0.00   & 0.01  & 0.00    & 0.00    & 0.00 & 0.02   & -  & -    & 0.51       & -        \bigstrut[b]\\
\hline
\multirow{3}[2]{*}{\shortstack{Part\\Motion}}     & PARIS\textsuperscript{\dag}~\cite{liu2023paris} & 0.00      & 0.00   & 0.03   & 0.00  & 0.00    & 0.00    & 0.03 & 0.08   & 0.06  & 0.00    & 0.77       & 0.31        \bigstrut[t]\\
                              & PARIS~\cite{liu2023paris}           & 0.04      & 0.00   & 0.05   & 0.04  & 0.03    & 34.81   & 0.03 & 0.18   & 0.27  & 0.30    & 1.48       & 1.16        \\
                              & PARIS*~\cite{liu2023paris}          & 0.04      & 0.00   & 0.00   & 0.03  & 0.00    & 0.72    & 0.00 & 0.10   & 0.28  & 0.30    & 1.57       & 0.58        \bigstrut[b]\\
\hline
\multirow{3}[2]{*}{CD-s}     & PARIS\textsuperscript{\dag}~\cite{liu2023paris} & 0.20      & 2.88   & 0.15   & 6.19  & 0.28    & 0.94    & 2.60 & 19.45  & 0.58  & 11.76   & 10.22      & 20.92       \bigstrut[t]\\
                              & PARIS~\cite{liu2023paris}           & 0.21      & 1.92   & 0.15   & 10.78 & 0.39    & 1.84    & 2.68 & 28.15  & 0.53  & 8.96    & 9.32       & 81.11       \\
                              & PARIS*~\cite{liu2023paris}          & 0.21      & 1.73   & 0.15   & 2.73  & 0.23    & 1.39    & 2.47 & 5.24   & 0.62  & 7.68    & 8.87       & 16.86      \bigstrut[b] \\
\hline
\multirow{3}[2]{*}{CD-m}     & PARIS\textsuperscript{\dag}~\cite{liu2023paris} & 0.53      & 1.13   & 0.14   & 0.43  & 0.23    & 0.85    & 0.89 & 0.27   & 5.13  & 20.67   & 67.54      & 101.20      \bigstrut[t]\\
                              & PARIS~\cite{liu2023paris}           & 0.55      & 1.43   & 0.14   & 7.09  & 0.25    & 2.49    & 0.83 & 0.24   & 7.16  & 83.54   & 77.48      & 143.17      \\
                              & PARIS*~\cite{liu2023paris}          & 0.41      & 1.21   & 0.15   & 0.76  & 0.20    & 0.97    & 0.62 & 0.26   & 6.63  & 49.14   & 92.26      & 12.89       \bigstrut[b]\\
\hline
\multirow{3}[2]{*}{CD-w}     & PARIS\textsuperscript{\dag}~\cite{liu2023paris}  & 0.42      & 2.68   & 0.25   & 6.07  & 0.30    & 0.96    & 1.80 & 18.31  & 0.46  & 8.12    & 8.20       & 18.98       \bigstrut[t] \\
                              & PARIS~\cite{liu2023paris}           & 0.43      & 1.95   & 0.25   & 9.24  & 0.33    & 1.76    & 1.97 & 30.30  & 0.43  & 7.63    & 7.84       & 68.77       \\
                              & PARIS*~\cite{liu2023paris}          & 0.37      & 1.81   & 0.26   & 2.68  & 0.26    & 1.10    & 1.64 & 4.97   & 0.44  & 7.40    & 6.43       & 15.75     \bigstrut[b] \\ 
\hline
\end{tabular}%
}
\vspace{-0.1in}
\caption{Results of the best-performing optimization trials from PARIS and PARIS* on PARIS Two-Part Dataset, where PARIS*~\cite{liu2023paris} is augmented with depth, PARIS\textsuperscript{\dag} are numbers reported in the original PARIS paper~\cite{liu2023paris} for reference. Note that Blade, Storage, and Real Storage have prismatic joints whose Axis Position Error is undefined.
}
\label{tab:supp_paris_best}
\end{table*}

\begin{table*}[h!]
\centering
\resizebox{0.98\textwidth}{!}{

\begin{tabular}{lcl|llllllllll}
\hline
    &                              &                         & {Axis Ang 0} & {Axis Ang 1} & {Axis Pos 0} & {Axis Pos 1} & {Part Motion 0} & {Part Motion 1} & {CD-s}       & {CD-m 0}     & {CD-m 1}     & {CD-w} \bigstrut[b]\\

\hline
\multirow{4}{*}{Fridge-m}  & \multirow{2}{*}{PARIS*-m~\cite{liu2023paris}} & mean\tsb{std} & 34.52\tsb{19.1} & 15.91\tsb{7.0} & 3.60\tsb{1.6}   & 1.63\tsb{1.3} & 86.21\tsb{55.2} & 105.86\tsb{43.6} & 8.52\tsb{2.0} & 526.20\tsb{141.6} & 160.86\tsb{102.2} & 15.00\tsb{3.5} \bigstrut[t]\\
                           &                                               & best trial    & 27.30           & 13.89          & 3.22            & 2.27          & 22.62           & 43.18            & 6.74          & 265.46            & 150.20            & 10.45          \bigstrut[b]\\ \cline{2-13}
                           & \multirow{2}{*}{Ours}                         & mean\tsb{std} & 0.16\tsb{0.0}   & 0.10\tsb{0.0}  & 0.01\tsb{0.0}   & 0.00\tsb{0.0} & 0.11\tsb{0.0}   & 0.13\tsb{0.0}    & 0.61\tsb{0.0} & 0.40\tsb{0.0}     & 0.52\tsb{0.0}     & 0.89\tsb{0.0}  \bigstrut[t]\\
                           &                                               & best trial    & 0.11            & 0.10           & 0.01            & 0.00          & 0.08            & 0.13             & 0.61          & 0.41              & 0.52              & 0.89           \bigstrut[b]\\
\hline
\multirow{4}{*}{Storage-m} & \multirow{2}{*}{PARIS*-m~\cite{liu2023paris}} & mean\tsb{std} & 43.26\tsb{25.1} & 26.18\tsb{7.2} & 10.42\tsb{19.1} & -             & 79.84\tsb{45.4} & 0.64\tsb{0.2}    & 8.56\tsb{1.1} & 128.62*     & 266.71\tsb{102.7} & 8.66\tsb{5.4}  \bigstrut[t]\\
                           &                                               & best trial    & 0.38            & 27.54          & 6.50            & -             & 47.06           & 0.91             & 9.15          & 128.62            & 216.96            & 21.95          \bigstrut[b]\\
                           \cline{2-13}
                           & \multirow{2}{*}{Ours}                         & mean\tsb{std} & 0.21\tsb{0.0}   & 0.88\tsb{0.2}  & 0.05\tsb{0.0}   & -             & 0.13\tsb{0.0}   & 0.00\tsb{0.0}    & 0.85\tsb{0.0} & 0.21\tsb{0.0}     & 3.46\tsb{2.8}     & 0.99\tsb{0.0}  \bigstrut[t]\\
                           &                                               & best trial    & 0.20            & 0.76           & 0.05            & -             & 0.13            & 0.00             & 0.85          & 0.21              & 0.23              & 0.99             \bigstrut[b] \\
\hline

\end{tabular}%

}
\caption{Additional result statistics on multi-part object dataset, including the average, standard deviation, and the best result across 10 trials with different random seeds.
PARIS*-m~\cite{liu2023paris} is augmented with depth and extended to handle objects with more than two parts. Joint 1 of ``Storage-m'' is prismatic and does not have Axis Position Error. PARIS*-m only reconstructed two non-empty parts in 9 out of 10 trials, and its CD-m 0 (chamfer distance of movable part 0) is reported from the only trial with three reconstructed parts.
}
\label{tab:supp_multi_part}
\vspace{-5pt}
\end{table*}

The original PARIS composites static field $\mathcal{F}^S$ and mobile field $\mathcal{F}^M$, each representing one object part. It also optimizes the axis and state change of the joint connecting the parts. For objects with $P$ parts, where $P > 2$, we extend PARIS to optimize $P$ fields, $\mathcal{F}^S, \mathcal{F}^{M_0}, \mathcal{F}^{M_1}, \ldots, \mathcal{F}^{M_{P - 2}}$, as well as $P - 1$ rigid transformations for each movable part. The per-point color composition in Eq. 1 of PARIS is directly extended to include $P$ terms, namely

\begin{align}
\hat{C}(\mathbf{r}) = \int_{h_n}^{h_f}(w^S(h)\cdot \mathbf{c}^S(h) + \sum_{i = 0}^{P - 2}w^{M_i}(h)\mathbf{c}^{M_i}(h))dh.
\end{align}

We also closely follow PARIS' fine-tuning procedure and use its proposed regularization loss $\mathcal{L}_{\text{prob}}$ that encourages each point to accumulate information only from one field. Formally,

\begin{align}
    \mathcal{L}_{\text{prob}} &= H(P_M(\mathbf{r})),\\
    P_M(\mathbf{r}) &= \frac{O^M(\mathbf{r})}{O^M(\mathbf{r}) + O^S(\mathbf{r})},\\
    H(x) &= -(x\cdot\log(x) + (1 - x)\cdot \log(1 - x)),
\end{align}

where $P_M(\mathbf{r})$ denotes the ratio of the contribution of the mobile field $\mathcal{F}^M$ to ray $\mathbf{r}$, $H$ is the binary entropy function.

For PARIS*-m, we extend $\mathcal{L}_{\text{prob}}$ to be the entropy over the P-ary probability distribution $(P_S, P_{M_0}, \ldots, P_{M_{P - 2}})$. Formally,
\begin{align}
    \mathcal{L}_{\text{prob}} & = H(P(\mathbf{r})) \\
    & = -\left(P_S(\mathbf{r})\log P_S(\mathbf{r}) + \sum_{i = 0}^{P - 2}P_{M_i}(\mathbf{r}) \log P_{M_i}(\mathbf{r})\right).
\end{align}

\section{Additional Results}

\subsection{Results on PARIS Two-Part Object Dataset}

Being optimization-based methods, PARIS, PARIS*, PARIS*-m, and our approach all have varying performances across trials depending on different initialization of the model parameters. For a comprehensive evaluation, we run each method 10 times with different random seeds (also set randomly) and report their mean and standard deviation in Table~\ref{tab:two_part} of the main paper. While our approach produces quite stable results across trials, PARIS and its variations have large performance variances. For completeness, in Table~\ref{tab:supp_paris_best} we summarize the results from the \textit{best} trials of PARIS and PARIS*, alongside the numbers reported in the original PARIS paper, denoted PARIS\textsuperscript{\dag} for reference. To select the best trial, we compute the minimum value for each metric across 10 trials, then choose the trial with the most number of minimum metric values. When there are ties, we prioritize the metrics with larger variances. 

For most objects, the best results from our re-run trials of PARIS are comparable to the reported numbers from PARIS\textsuperscript{\dag}. However, achieving such results takes many trials and drastic failure cases are not uncommon, as reflected by the overall large average errors. For challenging instances such as stapler, 10 trials are still insufficient to get one successful reconstruction. 

The best trials from depth-augmented PARIS* have comparable performance to PARIS on joint-related metrics, and overall better performance on part- and object-level chamfer distances, showing the usefulness of depth supervision. Nevertheless, as suggested in Table~\ref{tab:two_part} of the main paper, the introduction of one more loss term to PARIS' optimization process further destabilizes it and leads to larger variances and average errors.

\subsection{Results on Multi-Part Objects}

For multi-part objects, we follow the same protocol and run 10 trials with different random seeds for both PARIS*-m and our method. Their average, standard deviation, and best trials are summarized in Table~\ref{tab:supp_multi_part}. Our method continues to exhibit stable performance. On the other hand, PARIS*-m struggles to deal with the increased complexity of multiple movable parts. Even the best trials cannot successfully reconstruct both movable parts or their joint parameters.

\subsection{Additional Qualitative Results}

\begin{figure*}[]
\centering
\includegraphics[width=0.72\linewidth]{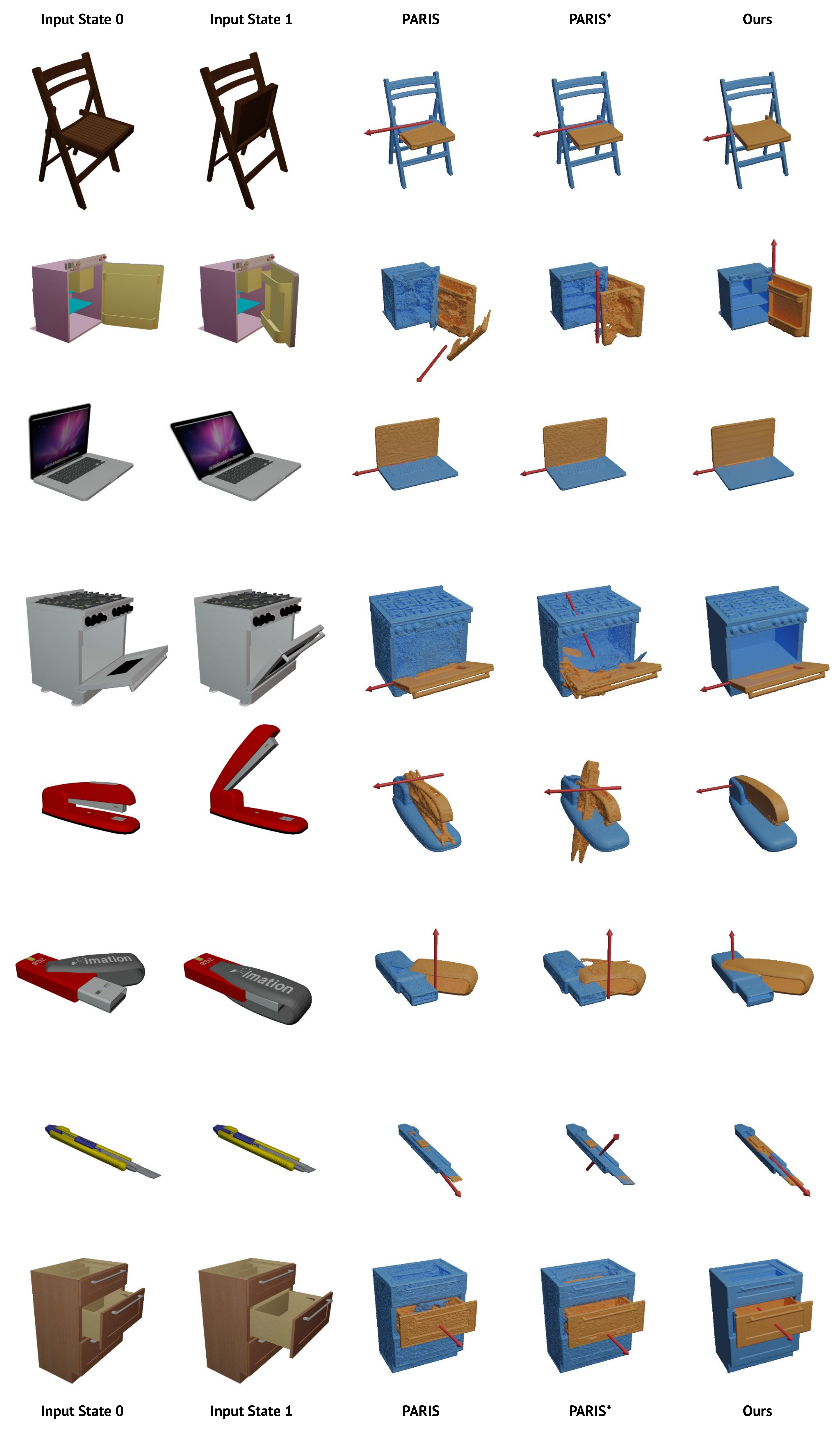}
\vspace{-5pt}
\caption{Additional visualizations of reconstruction results from PARIS, PARIS* (PARIS augmented with depth supervision), and our approach on synthetic objects from PARIS Two-Part Object Dataset. For each method, we selected typical trials with performance closest to the average performance. Please refer to our supplementary video \faVideoCamera~ for $360^{\circ}$ views and motion interpolation results.}
\label{fig:supp_qual_syn}
\end{figure*}

\begin{figure*}[h]
\centering
\includegraphics[width=0.95\linewidth]{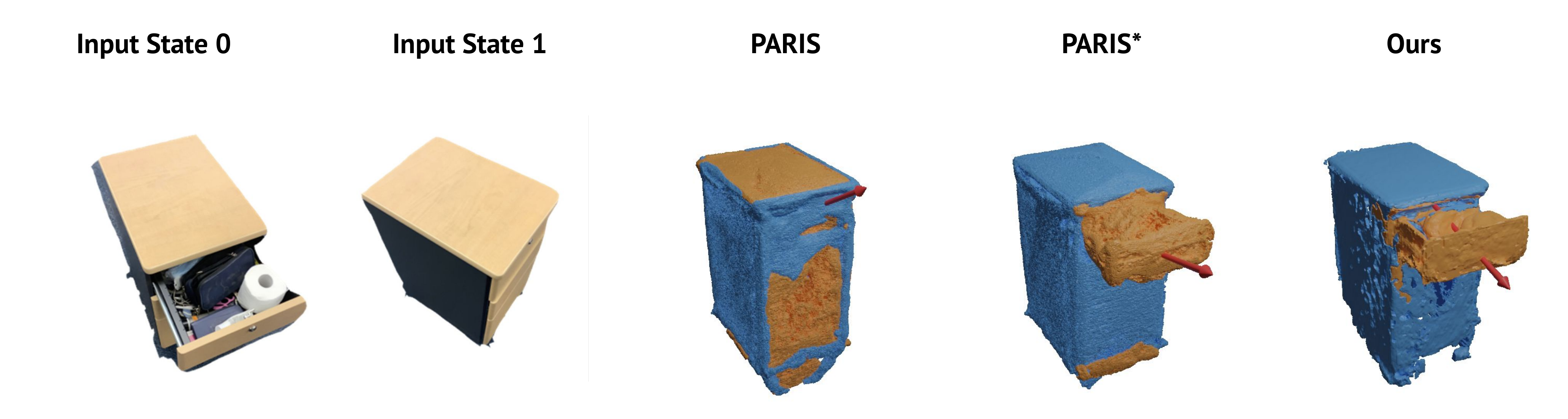}
\vspace{-5pt}
\caption{Additional visualizations of reconstruction results from PARIS, PARIS* (PARIS augmented with depth supervision), and our approach on the real storage furniture from PARIS Two-Part Object Dataset. For each method, we selected a typical trial with performance closest to the average performance. Please refer to our supplementary video \faVideoCamera~ for $360^{\circ}$ views and motion interpolation results.}
\label{fig:supp_qual_real}
\end{figure*}

Figures~\ref{fig:supp_qual_syn}, \ref{fig:supp_qual_real}~
show additional qualitative results from PARIS, PARIS*, 
and our method on PARIS two-part objects. We visualize the per-part reconstructions and reconstructed joint axes. Please refer to our supplementary video \faVideoCamera~for $360^{\circ}$ view of the reconstructions and motion interpolation results.

\subsection{Demo Interaction with the Reconstructed Multi-Part Storage in Simulation}

\begin{figure*}[h!]
    \begin{subfigure}{0.23\linewidth}
        \includegraphics[width=\linewidth]{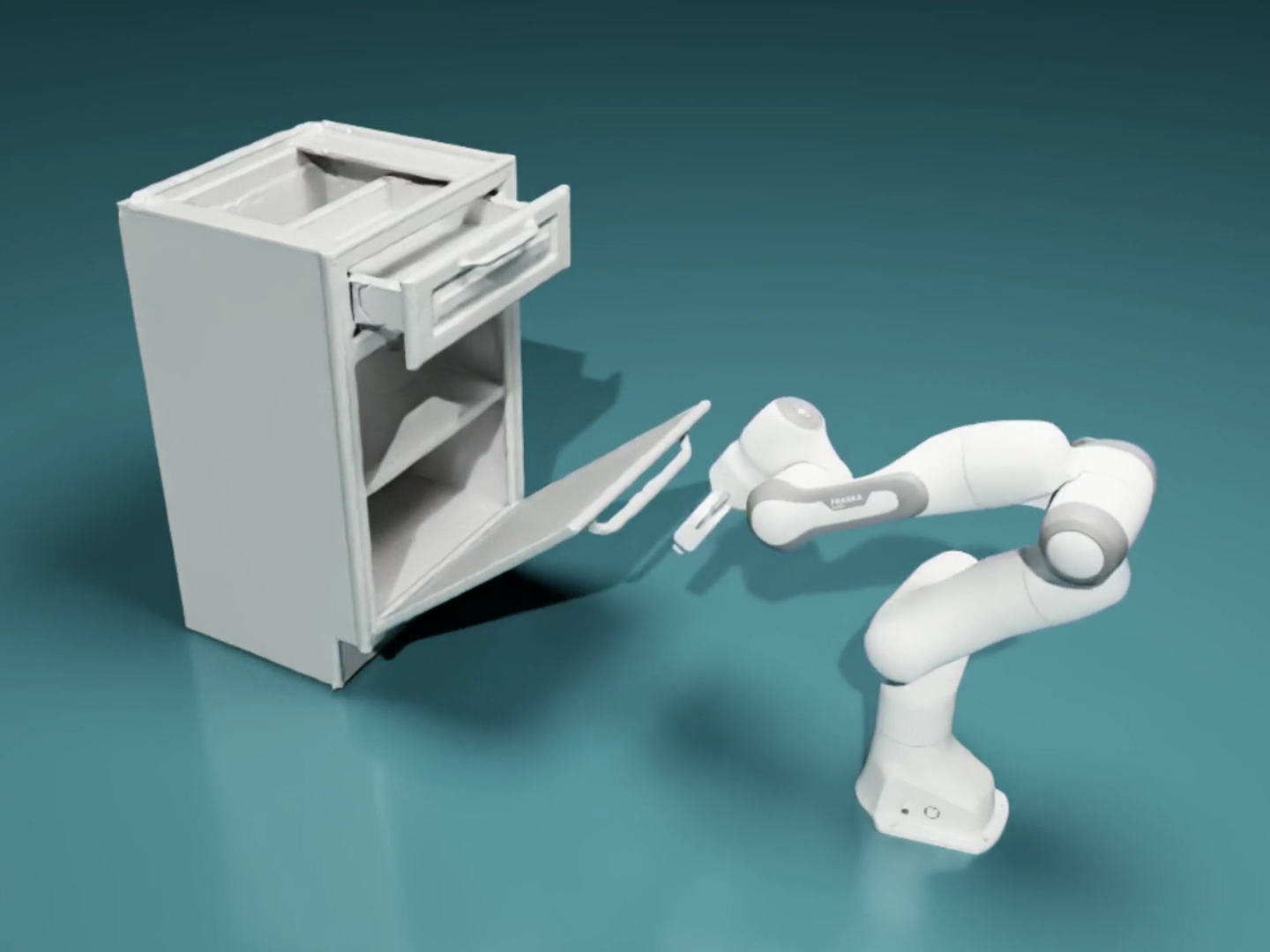}%
        \label{fig:demo_00}%
        \caption{}
    \end{subfigure}
    \hfill
    \begin{subfigure}{0.23\linewidth}
        \includegraphics[width=\linewidth]{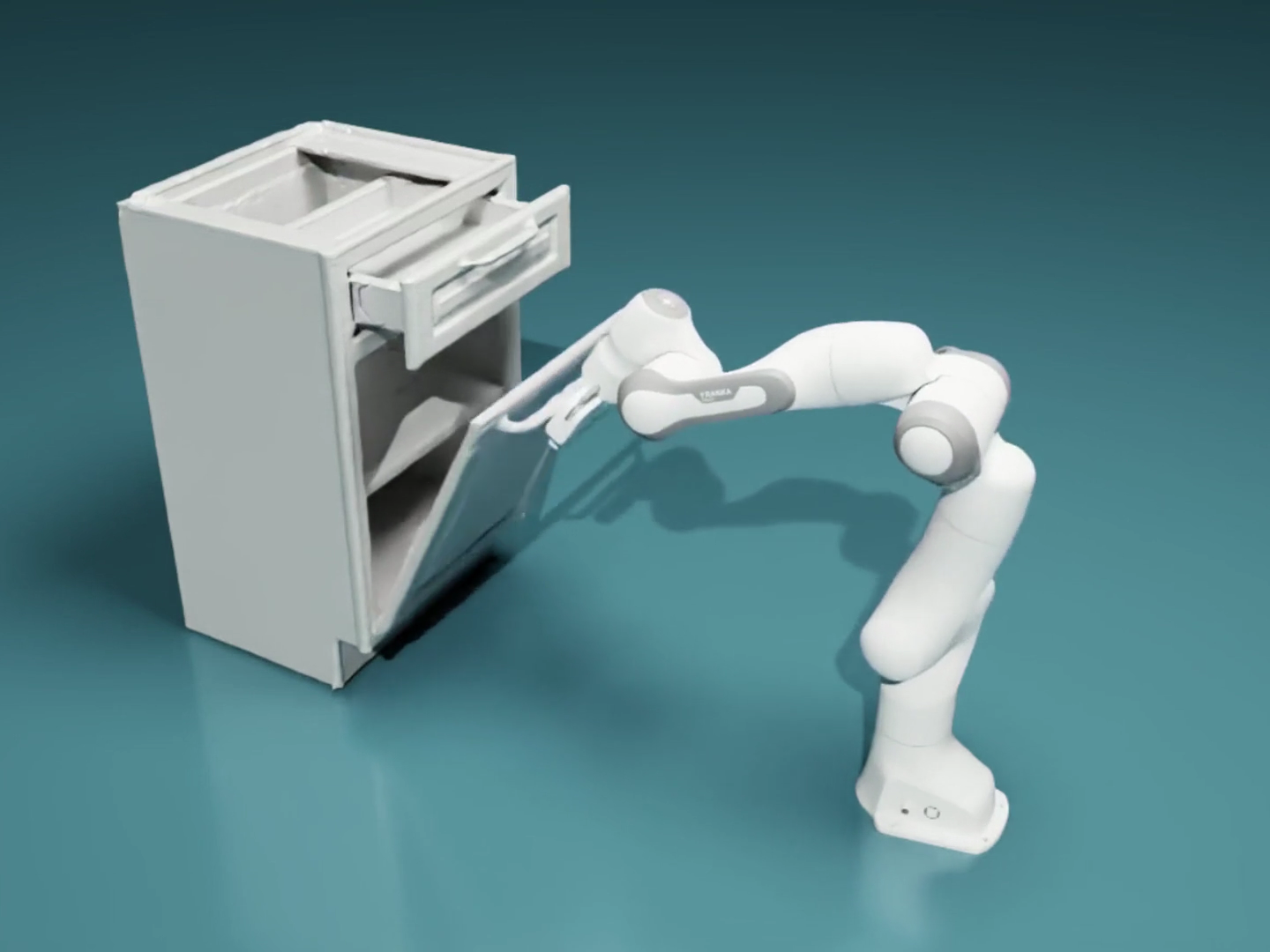}%
        \label{fig:demo_01}%
        \caption{}
    \end{subfigure}
    \hfill
    \begin{subfigure}{0.23\linewidth}
        \includegraphics[width=\linewidth]{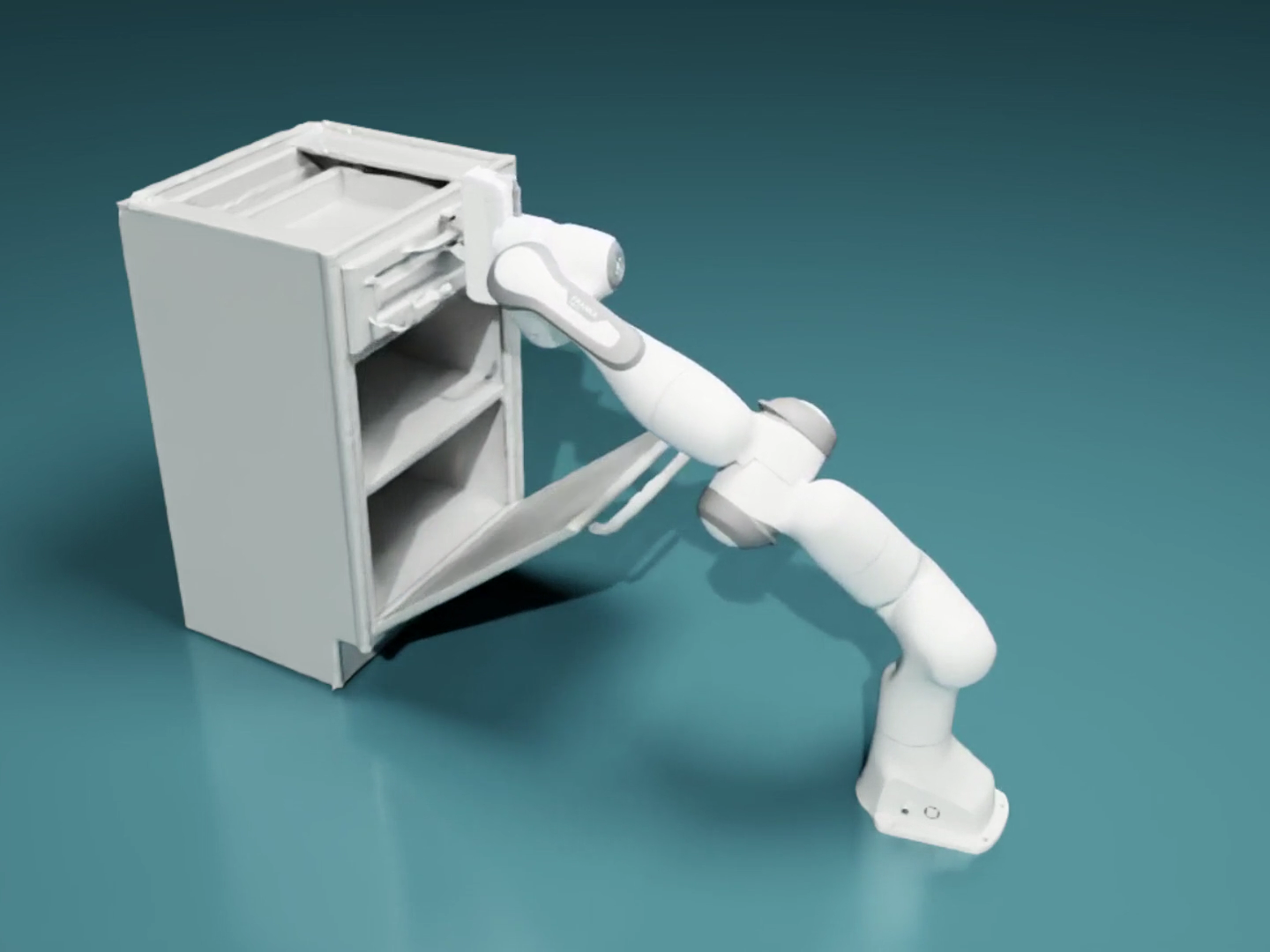}%
        \label{fig:demo_02}%
        \caption{}
    \end{subfigure}
    \hfill
    \begin{subfigure}{0.23\linewidth}
        \includegraphics[width=\linewidth]{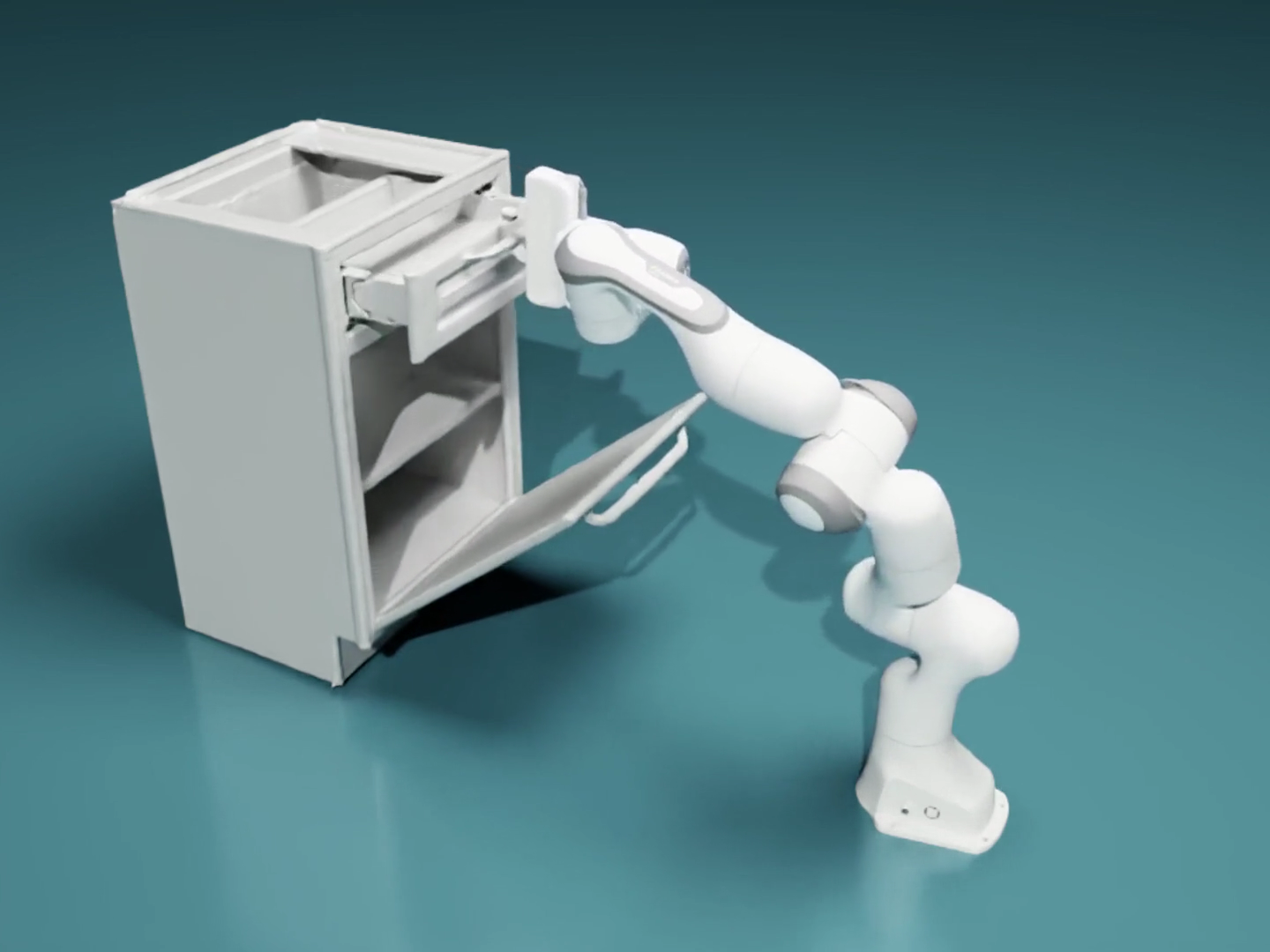}%
        \label{fig:demo_03}%
        \caption{}
    \end{subfigure}
    \caption{Screenshots from demo interaction sequence with our reconstructed storage furniture in Issac Gym. We controlled a Franka Emika robot arm to interact with both movable parts. Please refer to our supplementary video \faVideoCamera~ for the full sequence.}
    \label{fig:supp_robot_demo}

\end{figure*}
Our reconstructed digital twin can be readily imported to simulation environments and interacted with. Figure~\ref{fig:supp_robot_demo} shows screenshots of a demo interaction sequence we made. Please refer to our supplementary video \faVideoCamera~ for the full sequence. We imported our reconstruction of the multi-part storage furniture (``Storage-m'') into Issac Gym, using both the reconstructed meshes and the joint parameters. We generated a control sequence for a Franka Emika robotic arm to interact with both movable parts of the reconstruction. As illustrated by the demo, the ability to reconstruct digital twins of multi-part articulated objects enables exciting real2sim applications.

\end{document}